%% file: main.tex
\pdfoutput = 1

\documentclass{article}
\usepackage{ijcai21}
\usepackage{times}
\usepackage{soul}
\usepackage{url}
\usepackage[hidelinks]{hyperref}
\usepackage[utf8]{inputenc}
\usepackage[small]{caption}
\usepackage[pdftex]{graphicx}
\usepackage{wrapfig}
\usepackage[dvipsnames]{xcolor}
\usepackage{amsmath}
\usepackage{amsthm}
\usepackage{booktabs}
\usepackage{algorithm}
\usepackage{algorithmic}
\usepackage{color,soul}
\usepackage{subfig}
\usepackage[title]{appendix}

\title{Discrepancies in Epidemiological Modeling of Aggregated Heterogeneous Data}

\author{
Anna L. Trella$^{1}$\and
Peniel N. Argaw$^{1,}$\footnote{Equal contribution}\and
Michelle M. Li$^{2,}$\footnotemark[1]\and
James A. Hay$^3$
\affiliations
$^1$Harvard John A. Paulson School Of Engineering and Applied Sciences, $^2$Harvard Medical School, $^3$Harvard T.H. Chan School of Public Health
\emails
\{annatrella, peniel, michelleli\}@g.harvard.edu, jhay@hsph.harvard.edu
}

\begin{document}

\maketitle

\begin{abstract}
\input{000abstract}
\end{abstract}

\section{Introduction}
\input{010intro}

\section{Methods}
\label{sec:methods}
\input{020methods}

\section{Simulated Scenarios}
\label{sec:scenarios}

\input{030scenarios}

\section{Results}
\label{sec:results}
\input{040results}

\section{Conclusion}
\input{050conclusion}

\section{Broader Impact Statement}
\input{060impact}

\section*{Acknowledgements}
We are grateful for the guidance and support from Dr.~Michael Mina and Dr.~Milind Tambe throughout our project. M.M.L.~is supported by T32HG002295 from the National Human Genome Research Institute and a National Science Foundation Graduate Research Fellowship.

\bibliographystyle{named}
\bibliography{reference}

\input{070appendix}

\end{document}

%% file: 000abstract.tex
Within epidemiological modeling, the majority of analyses assume a single epidemic process for generating ground-truth data. However, this assumed data generation process can be unrealistic, since data sources for epidemics are often aggregated across geographic regions and communities. As a result, state-of-the-art models for estimating epidemiological parameters, e.g.~transmission rates, can be inappropriate when faced with complex systems. Our work empirically demonstrates some limitations of applying epidemiological models to aggregated datasets. We generate three complex outbreak scenarios by combining incidence curves from multiple epidemics that are independently simulated via SEIR models with different sets of parameters. Using these scenarios, we assess the robustness of a state-of-the-art Bayesian inference method that estimates the epidemic trajectory from viral load surveillance data. We evaluate two data-generating models within this Bayesian inference framework: a simple exponential growth model and a highly flexible Gaussian process prior model. Our results show that both models generate accurate transmission rate estimates for the combined incidence curve at the cost of generating biased estimates for each underlying epidemic, reflecting highly heterogeneous underlying population dynamics. The exponential growth model, while interpretable, is unable to capture the complexity of the underlying epidemics. With sufficient surveillance data, the Gaussian process prior model captures the shape of complex trajectories, but is imprecise for periods of low data coverage. Thus, our results highlight the potential pitfalls of neglecting complexity and heterogeneity in the data generation process, which can mask underlying location- and population-specific epidemic dynamics.

{\let\thefootnote\relax\footnote{Our Python implementation of the Bayesian inference method using viral load surveillance data is open source and available at \href{https://github.com/pargaw/pyCt}{\texttt{https://github.com/pargaw/pyCt}}}}

%% file: 010intro.tex
Reliable testing and reporting of new cases play key roles in early detection and management of an epidemic. For example, testing data are often used to generate predictions regarding the dynamics of an epidemic \cite{tolles2020modeling} to guide interventions enforced by governments \cite{brauner2021inferring}. Data sources for epidemics are usually aggregated across geographic regions and spatial scales \cite{dong2020interactive}. However, aggregating datasets can mask spatial heterogeneity in underlying incidence trends \cite{chang2021mobility}. In fact, it has been shown that individual-level data can better inform risk models' predictions \cite{xu2020epidemiological}, especially for disadvantaged communities \cite{chang2021mobility}.

Still, due to limited access to data (e.g. during early detection stages), epidemiologists must simulate epidemics to generate scenarios or predictions for possible outcomes of interventions \cite{tolles2020modeling}. Because of the complexities in disease and population dynamics, routinely used epidemiological models (e.g.~compartmental models) make assumptions that can be oversimplifying \cite{tolles2020modeling}.

For the scope of this work, we adopt the recent \textit{Ct model} to evaluate our simulated complex outbreak scenarios. The Ct model estimates epidemiological dynamics from viral load distributions, namely cycle threshold (Ct) values from real-time PCR assays \cite{hay2020estimating}. Ct values represent the viral load in a person at a given point in time, and the distribution of detectable viral loads under random cross-sectional surveillance has been shown to be informative of the epidemic growth rate \cite{hay2020estimating}. Specifically, low Ct values ($\text{Ct} \leq 29$) correspond to high viral loads observed, indicating recent epidemic growth, while high Ct values ($38 \leq \text{Ct} \leq 40$; Ct values above 40 cannot be measured due to technical limitations of real-time PCR assays) correspond to low viral loads observed, indicating recent epidemic decline \cite{hay2020estimating}. Still, the Ct model, like other state-of-the-art (SoTA) models for estimating epidemiological dynamics, assumes a single epidemic process that generates the data. We use the Ct model as a demonstration---any epidemiological model can be applied instead---to evaluate an epidemiological model's robustness in differentiating complex population structures and epidemiological dynamics.

Our evaluation process comprises of two main components: simulating three scenarios in which two distinct populations---with different susceptibility, infectious, incubation, and recovery rates---interact (Sections \ref{sec:methods} and \ref{sec:scenarios}); and measuring the fit of the Ct model on the combined populations' viral load distributions (Sections \ref{sec:methods} and \ref{sec:results}). We also empirically demonstrate the model's sensitivity to the choice of a posterior sampling method (Section \ref{sec:results}).

Ultimately, our results highlight some of the challenges in implementing and deploying epidemiological models. While it is clearly naive to assume a single data-generating process for a dataset constructed by two underlying epidemic processes, we hypothesize that estimates from an epidemiological model can only reflect one of the epidemics and/or populations, depending on the stage of the two underlying epidemics and how well the data represent the two populations. For example, suppose we have an abundance of surveillance data from population \#1 that indicate an epidemic decline, and sparse surveillance data from population \#2, which is entering an epidemic growth phase. After combining the datasets, we must be careful not to incorrectly conclude that population \#2 is also in the decline phase of the epidemic. Blindly aggregating data can lead to flawed policy decisions. Thus, as shown through this work, our recommendation is to evaluate the level of aggregation in the data before drawing conclusions (e.g. Given only state-aggregated data, how well-equipped are we to make county-level decisions?).

%% file: 020methods.tex
We begin with an overview of the models used to simulate and evaluate the outbreak scenarios described in Section \ref{sec:scenarios}. We utilize these simulated data to assess the accuracy of the inference approach in estimating the true epidemic trajectory. We note that our simulation settings are set to broadly reflect a SARS-CoV-2 outbreak, but our findings can be considered pathogen-nonspecific.

\subsection{Simulations via SEIR model}
We simulate our complex scenarios with a compartmental model, specifically the Susceptible-Exposed-Infectious-Recovered (SEIR) model. We use a basic SEIR model \cite{ma2020estimating}, where the population size $N = S + E + I + R$, and the transition rates between the compartments $S$, $E$, $I$, and $R$ are defined by
\begin{align}
    \frac{dS}{dt} &= -\beta SI \\
    \frac{dE}{dt} &= \beta SI - \sigma E\\
    \frac{dI}{dt} &= \sigma E - \gamma I \\
    \frac{dI}{dt} &= \gamma I
\end{align}
where $\beta$ is the transmission rate (or rate in which susceptible individuals get exposed), $\sigma$ is the incubation rate (or rate at which exposed individuals become infectious), and $\gamma$ is the recovery rate (or rate at which infectious individuals recover).

\begin{table}
    \caption{\textbf{Default parameters for the simulated scenarios.} We set the following default values for the parameters in our SEIR model.}
    \label{tab:default}
    \centering
    \begin{tabular}{cc}
    \toprule
    Parameter & Value\\
    \midrule
    $N$ & 1,000,000 \\
    $R_0$ & 2.5 \\
    $\gamma$ & 1/8 \\
    $\sigma$ & 1/4 \\
    $S_{initial}$ & $0.9999N$ \\
    $E_{initial}$ & 0 \\
    $I_{initial}$ & $0.0001N$ \\
    $R_{initial}$ & 0 \\
    \bottomrule
    \end{tabular}
\end{table}

\begin{table}
    \caption{\textbf{Scaling factors for the simulated outbreak scenarios.} For our simulations, we define scaling factors for the SEIR model parameters $N$, $R_0$, and $\gamma$, where $R_0 = \beta / \gamma$. We only report the scaling factors for population~\#1 because the parameters are scaled relative to population~\#2, whose parameters are scaled by 1.}
    \centering
    \label{tab:parameters}
    \begin{tabular}{lccc}
    \toprule
    Scenario & $N$ & $R_0$ & $\gamma$\\
    \midrule
    Out-of-state travel (Section \ref{scenario:travel}) & 1 & 2 & 1.5 \\
    Seasonal travel (Section \ref{scenario:seasonal}) & 0.5 & 0.8 & 0.8 \\
    NPIs (Section \ref{scenario:npi}) & 1 & 0.8 & 1 \\
    \bottomrule
    \end{tabular}
\end{table}

\subsection{Simulations via Ct model}
After generating epidemic incidence curves for each of the scenarios using the SEIR model, we simulate observable Ct values under random cross-sectional surveillance, assuming that infected individuals are tested at an unknown point in their infection. The frequency and number of tests simulated are specified in Table \ref{tab:ct_values}. To capture the fact that measured Ct values depend on the age of an infection, $a$, we use a previously described model for modal viral load (or Ct) kinetics, $C_{mode}(a)$, assuming that observed Ct values follow a Gumbel distribution \cite{hay2020estimating}
\begin{equation}
    C(a) \sim Gumbel(C_{mode}(a), \sigma(a))
\end{equation}
where $\sigma(a)$ captures the observation that the Ct values exhibit less variability in older infections. We also assume that the parameters underlying the Ct model are known precisely based on previous estimates \cite{hay2020estimating}.

\subsection{Single Cross-Section Model}
We infer epidemic growth rates by fitting an exponential growth model given a single cross-section of observed Ct values. Briefly, the aim is to infer the recent growth rate of the epidemic using the distribution of Ct values obtained at a single point in time. An exponential growth model is simple yet interpretable because it has only one free parameter, which denotes the direction and magnitude of the epidemic trajectory from a single cross-section of data. We assume that the daily incidence over the $A_{max}$ days before the testing day $t$ grows or declines exponentially with rate $\beta$ such that
\begin{equation}
    \pi_{t - a} = \pi_0 e^{\beta(t - a)}
\end{equation}
where $\beta$ is the logarithm of the daily growth rate in incidence over the days $t - A_{max}$ and $t - 1$, $\pi_0$ is a nuisance parameter that can be ignored, and we assume a prior $\beta \sim \mathcal{N}(0, 0.25)$.

\subsection{Multiple Cross-Section Model}
We also infer epidemic growth rates by fitting a Gaussian process model given multiple cross-sections. Gaussian Processes (GPs) \cite{rasmussen2003gaussian} are a powerful non-parametric framework for performing Bayesian inference. The framework pairs a GP prior on daily incidence rate \cite{xu2016bayesian} with a custom Ct likelihood function. The flexibility of the GP prior allows for practitioners to analyze multiple test days, as the priors do not assume specific parameters. Our choices of a covariance function and hyperparameters are manually specified by domain experts. \cite{hay2020estimating} uses a modified version of the squared exponential (SE) kernel:
\begin{equation}
    k(x, x') = \eta^2 \exp \left[-\rho^2(x - x')^2\right]
\end{equation}
with hyperparameters $\eta = 1.5, \rho = 0.03.$ Therefore, daily incidence is defined as
\begin{equation}
    \pi_{t - a} = (1 + \exp(-f))^{-1}
\end{equation}
where $f \sim \mathcal{N}(\mathbf{0}, \mathbf{K})$ and $\mathbf{K}$ is a covariance matrix generated by $k$. For sampling, we use Hamilton Monte Carlo~\cite{duane1987hybrid} with a No-U-Turn sampler \cite{hoffman2014no} because it is a more efficient MCMC method than Metropolis-Hastings \cite{robert1999metropolis}, the sampler used in the implementation by \cite{hay2020estimating}.

%% file: 030scenarios.tex
\begin{figure*}[ht]
    \centering
    \subfloat(a){\includegraphics[width=0.95\columnwidth]{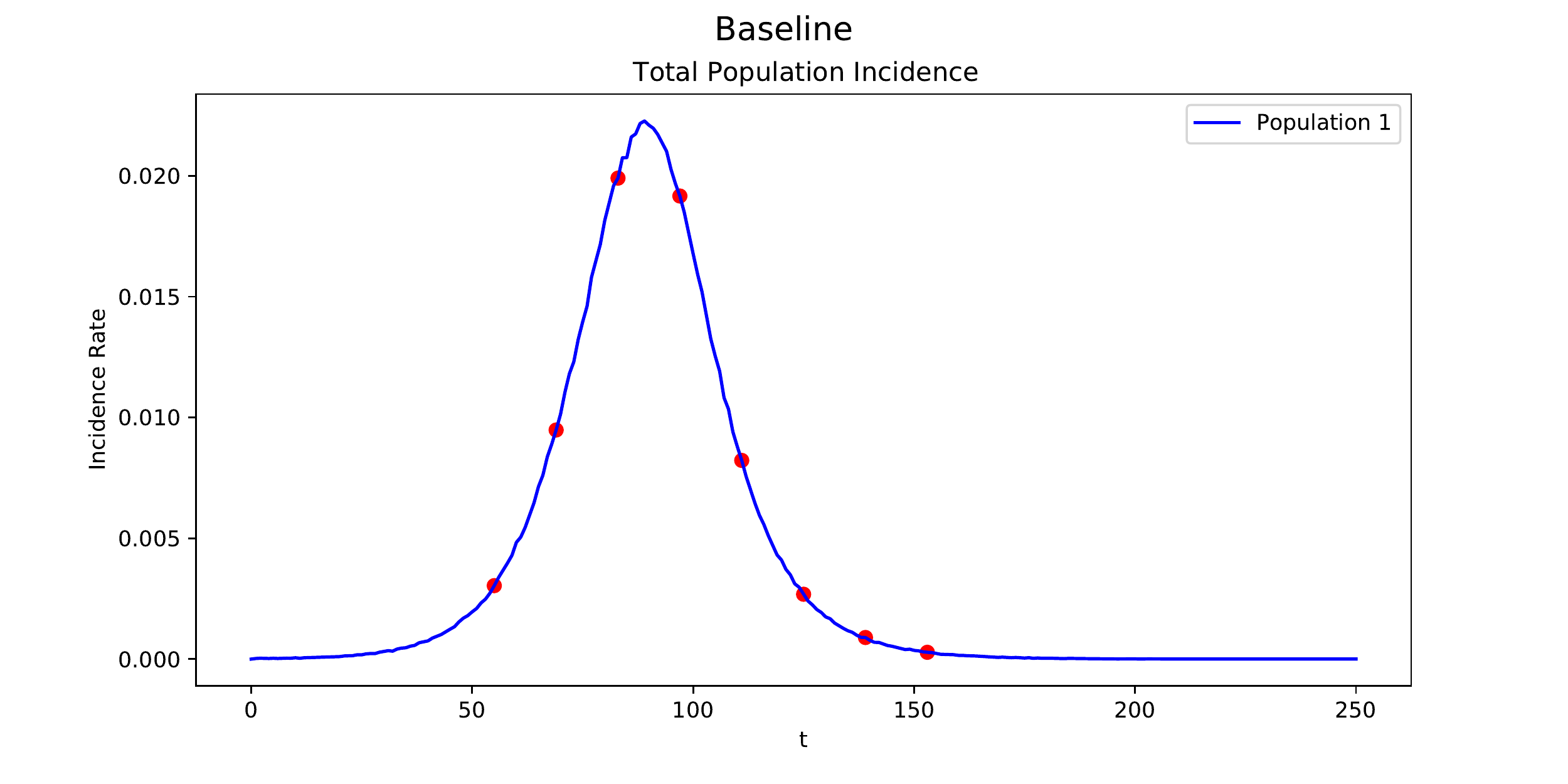}}
    \subfloat(b){\includegraphics[width=0.95\columnwidth]{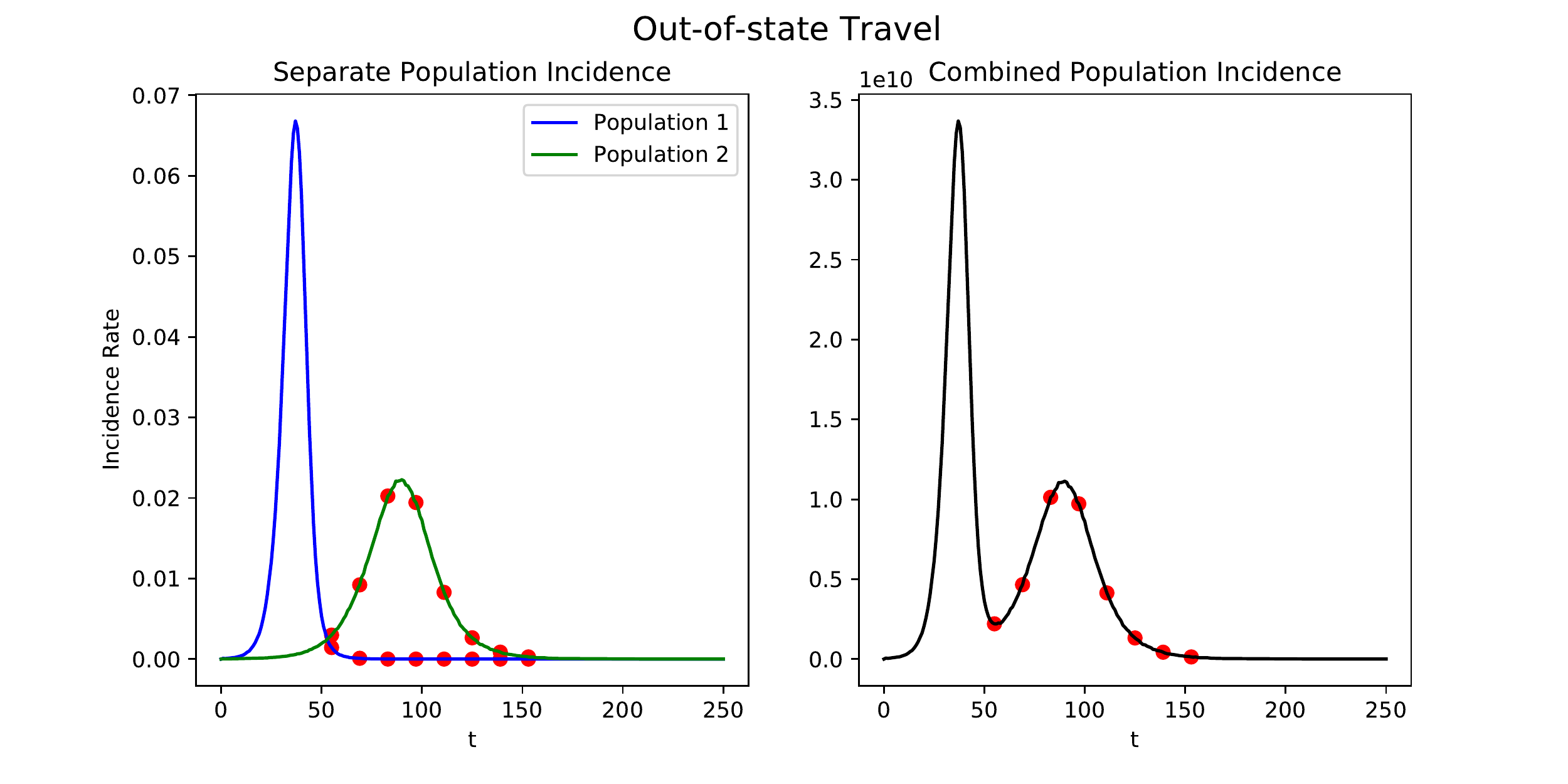}}
    \subfloat(c){\includegraphics[width=0.95\columnwidth]{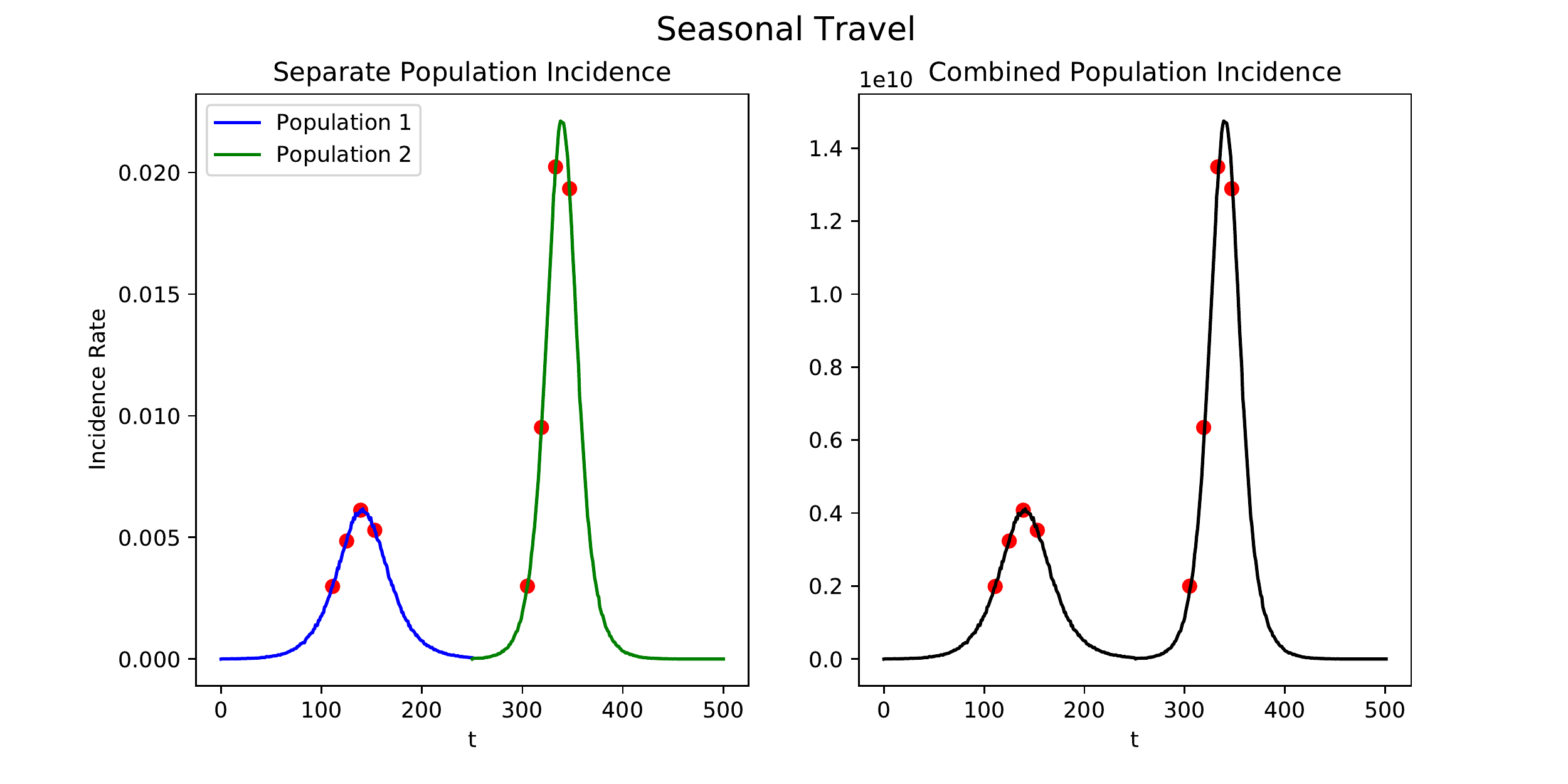}}
    \subfloat(d){\includegraphics[width=0.95\columnwidth]{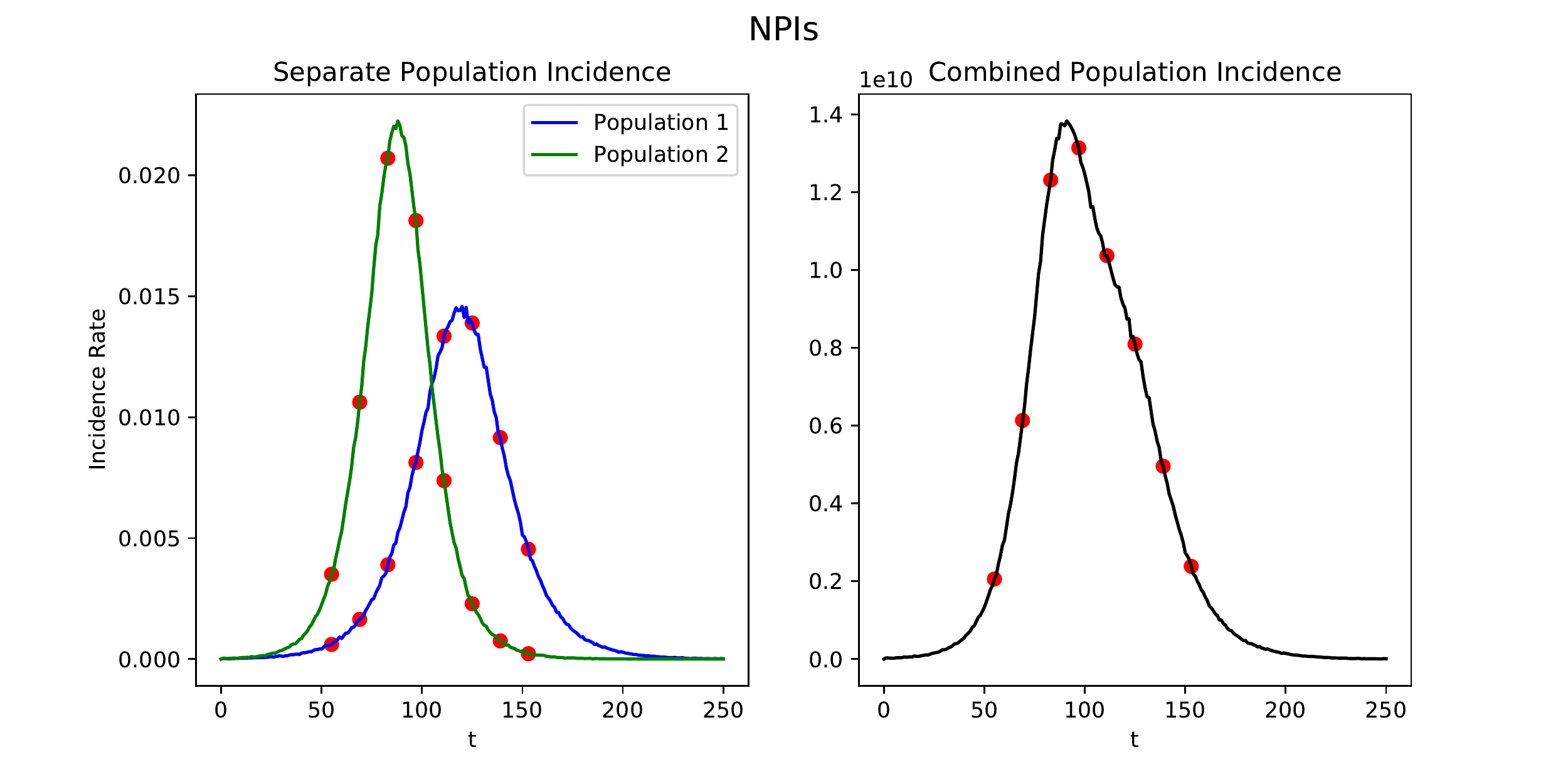}}
    \caption{\textbf{Incidence rates across individual and aggregated populations for each outbreak scenario.} \textbf{(a)} The baseline scenario consists of only one population (Section \ref{scenario:baseline}), whose incidence rates are shown in blue. Each of the \textbf{(b)} out-of-state travel (Section \ref{scenario:travel}), \textbf{(c)} seasonal travel (Section \ref{scenario:seasonal}), and \textbf{(d)} NPIs (Section \ref{scenario:npi}) scenarios has two panels: on the left are two incidence rate curves from populations \#1 (blue) and \#2 (green); and on the right is the incidence rate curve of the aggregated dataset. The red dots indicate times when testing is performed.}
    \label{figs/all_scenario_inc}
\end{figure*}

We next introduce our scenarios for evaluating the robustness of the Ct model. We create the scenarios by independently simulating outbreaks in two populations, and then combining their simulated Ct values observed through surveillance testing (Figure~\ref{figs/sim_ct_violin}). In each scenario, we arbitrarily define scaling factors for SEIR model parameters (Table~\ref{tab:parameters}) to generate outbreaks with different transmission dynamics (Figure~\ref{figs/all_scenario_inc}). The choices of parameters and scaling factors are designed to qualitatively demonstrate the desired epidemiological dynamics within the following scenarios, and we note that other parameter choices may give different dynamics but qualitatively similar results.

\subsection{Baseline} \label{scenario:baseline}

As a baseline, we simulate a simple scenario with only one population using the default parameters in Tables \ref{tab:default} and \ref{tab:ct_values}.

\subsection{Out-of-State Travel} \label{scenario:travel}

We simulate a scenario in which population \#1 consists of younger individuals (e.g. college students) who travel into a community (population \#2) from out-of-state (e.g. for spring break). We assume that the rate of recovery ($\gamma$) in the incoming younger individuals is higher on average than that of the community, broadly reflecting the epidemiological situation for SARS-CoV-2, where younger individuals are less likely to experience symptomatic infections \cite{davies2020age}. We also assume that the potential for transmission ($R_0$) is higher in population \#1 to reflect higher numbers and intensity of contacts amongst younger individuals. Further, the Ct model is built to incorporate variation across a population, including in age structures \cite{hay2020estimating}. Here, we assume that individuals in the two populations follow the same viral kinetics model (Table \ref{tab:ct_values}), and only vary the SEIR model parameters in Table \ref{tab:parameters}. We combine the populations by adding their Ct values at corresponding times $t \in [0, 250]$ (Figure \ref{figs/all_scenario_inc}).

\subsection{Seasonal Travel} \label{scenario:seasonal}

We simulate the dynamics of an epidemic in a community (e.g. college town) as a result of seasonal travel (e.g. students in town for the semester or leaving during the breaks). Population \#1 consists of individuals found in the town during the breaks; population \#2 consists of those present during the semester. In this scenario, the effect of population density can be modeled using a density-dependent compartment model (Table \ref{tab:parameters}). The transmission rate is defined as $\beta = pC$, where $p$ is the probability of infection per contact, and $C$ is the number of contacts per unit time. We assume that population \#1 has a lower transmission potential since the number of contacts per unit time decreases ($R_0$ is scaled by $0.8$). As the younger population leaves the town, population size and recovery rate also decrease. We combine the two populations by specifying the outbreaks in populations \#1 and \#2 to occur at $t \in [0, 250]$ and $t \in [251, 500]$, respectively (Figure \ref{figs/all_scenario_inc}).

\subsection{Non-Pharmaceutical Interventions} \label{scenario:npi}

We simulate a scenario in which non-pharmaceutical interventions (NPIs) (e.g. curfews, mask mandates) are enforced in one population (population \#1) but not in another (population \#2). We assume that the enforcement of NPIs results in a lower transmission rate ($R_0$) (Table \ref{tab:parameters}). We combine the populations as described in Section \ref{scenario:travel} (Figure \ref{figs/all_scenario_inc}).

%% file: 040results.tex
The exponential growth and GP prior models output average growth rates for a set of 8 defined time points (or 16 in the seasonal travel scenario by the exponential growth model). At each time point $t$, we show the predicted growth rate (namely, $\beta$ for the exponential model and $f$ for the GP prior model) and the true growth rate derived from the true incidence values. To evaluate the robustness of the two models, we compare the predicted growth rates against the true growth rates within a 95\% credible interval. Predictions are generated using only observed Ct values as input data (i.e.~Ct~$< 40$), so only data from infected individuals are included in the inference. Thus, estimates for the epidemic trajectory give relative, rather than absolute, incidence rates since we do not include information on the proportion of individuals who are infected.

\subsection{Baseline}

\begin{figure*}
    \centering
    {\includegraphics[width=\textwidth]{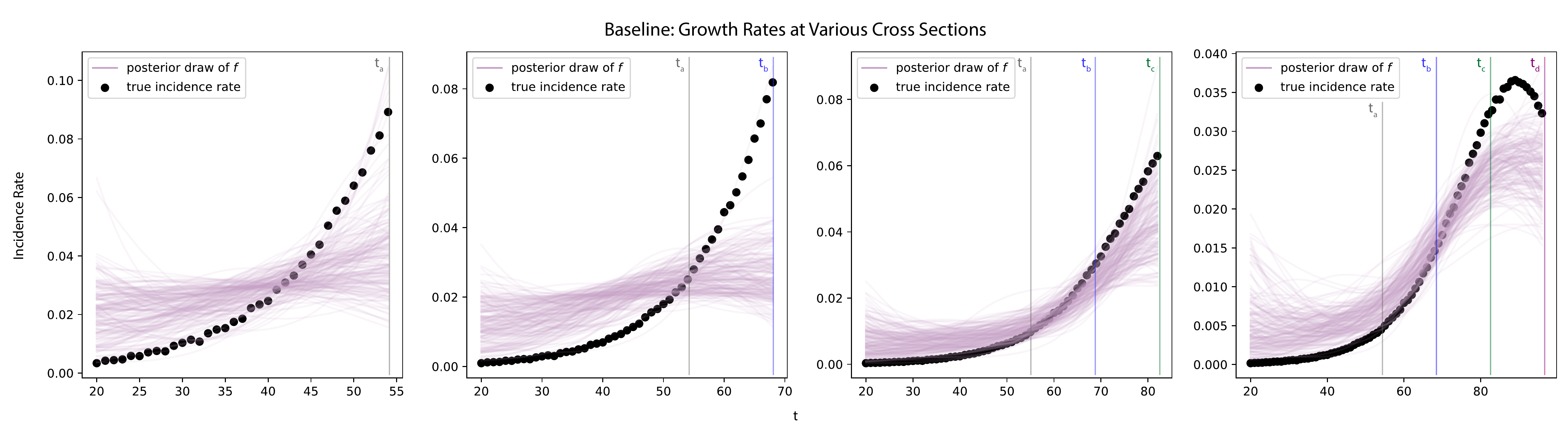}}
    \caption{\textbf{Limitations of the Gaussian process prior model.} We use the baseline scenario to demonstrate the limitations of the GP prior model. As the number of cross-sections increase (the leftmost panel shows one cross-section; the rightmost panel, four), the GP prior model's ability to capture the trajectory of the epidemic improves.}
    \label{figs/gp_limitations}
\end{figure*}

We first compare both methods' performance on a single standard incidence rate trajectory generated from a SEIR model (Section \ref{scenario:baseline}). Both models are able to estimate the correct direction of the epidemic, and the GP prior model can even capture the full shape of the trajectory as we increase the number of cross-sections (Figures \ref{figs/all_exp_results}a and \ref{figs/gauss_results}a). However, despite the simple trajectory, the exponential growth model has difficulty predicting growth rates near the inflection point of the curve (Figure~\ref{figs/scenario0_exp_results}), and the GP prior model needs---in our case---at least two cross-sections to capture the shape (Figure~\ref{figs/gp_limitations}).

\subsection{Out-of-State Travel} \label{results:travel}

In the out-of-state travel scenario, our exponential model is able to fit the combined incidence curve within all testing times except for when there is high variability across the two populations (namely, $t \in \{55, 69\}$) (Figure \ref{figs/scenario1_exp_results}). This is due to the fact that the aggregated dataset's incidence rate is the mean of the two populations' incidence rates (Figure~\ref{figs/all_scenario_inc}a). Since the rate of decrease in population~\#1 is greater than the rate of increase in population~\#2 at $t \in \{55, 69\}$, the predicted growth rate shows a sharper decline than the true growth rate (Figures \ref{figs/all_exp_results}b and \ref{figs/scenario1_exp_results}). Interestingly, the mean predicted growth rate for $t = 83$ is approximately two times greater than the true growth rate, although it is still within the 95\% credible interval (Figure \ref{figs/all_exp_results}b).

The GP prior model is able to fully capture the shape of the trajectory of population \#2, but not that of the trajectory of population \#1 (Figure \ref{figs/gauss_results}b). The peak of population \#1's epidemic occurs before the first testing time point (Figure \ref{figs/all_scenario_inc}b), so the GP prior model is only able to infer the epidemic decline in population \#1 (Figure \ref{figs/gauss_results}b). As a result of this limitation in surveillance data, even the GP prior model cannot accurately fit the initial incidence curve of population \#1 despite its overconfident uncertainty intervals (Figure \ref{figs/gauss_results}b). 

\subsection{Seasonal Travel} \label{results:seasonal}
As there is no overlap between the two populations in the seasonal travel scenario (Figure \ref{figs/all_scenario_inc}b), there is very little variability in the aggregated dataset. Thus, the exponential model seems to be robust in most of the sampled times (Figures \ref{figs/all_exp_results}c and \ref{figs/scenario2_exp_results}). The GP prior model is able to capture the full trajectory, including both peaks, as well because their testing time points are densely scattered (Figure \ref{figs/gauss_results}c). We also note that the GP prior model is able to handle the aleatoric uncertainty well, as shown by the large uncertainty intervals between times $t \in [175, 250]$, when no surveillance data is available.

\subsection{Non-Pharmaceutical Interventions} \label{results:npi}

\begin{figure*}
    \centering
    \subfloat(a){\includegraphics[width=0.95\columnwidth]{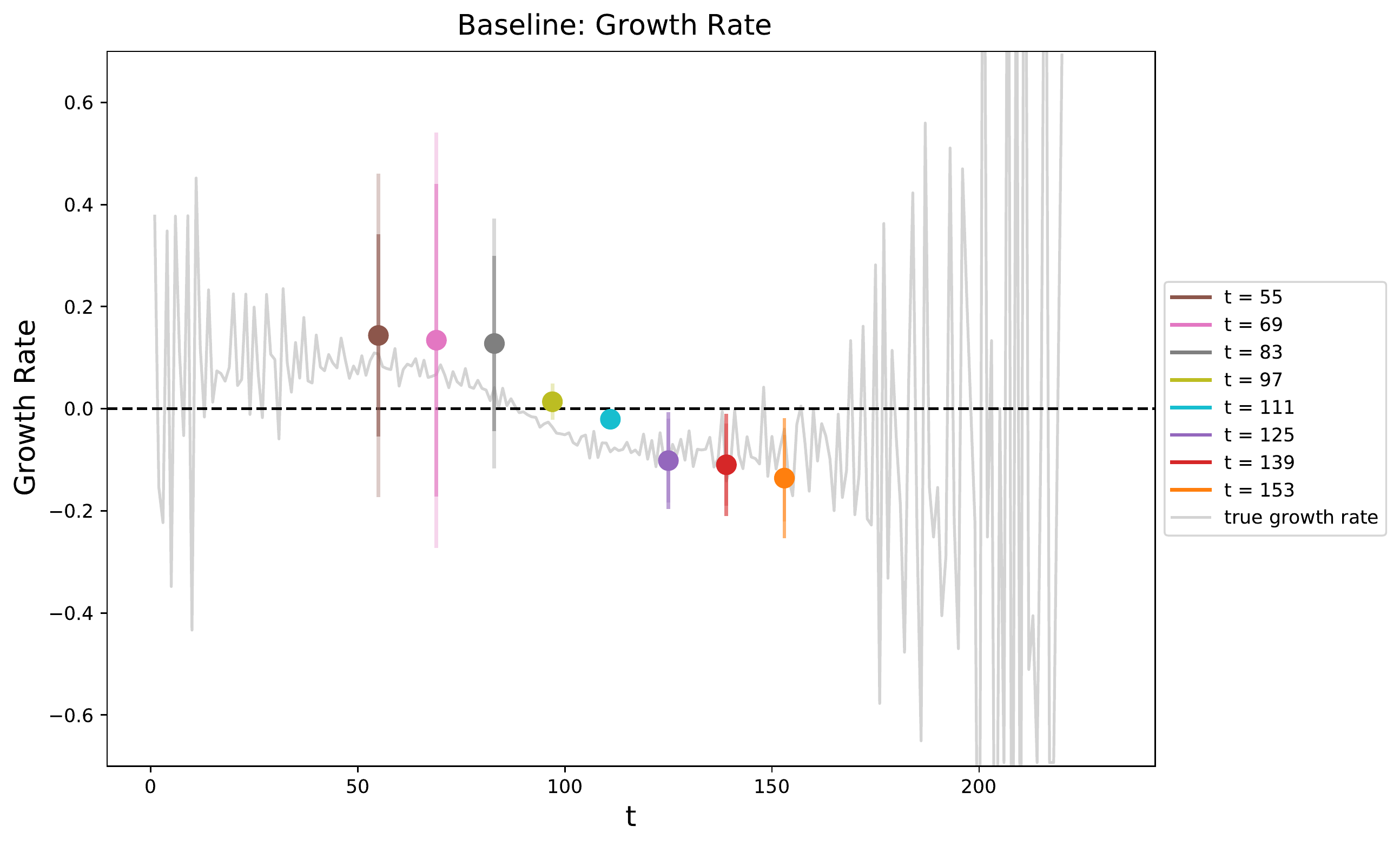}}
    \subfloat(b){\includegraphics[width=0.95\columnwidth]{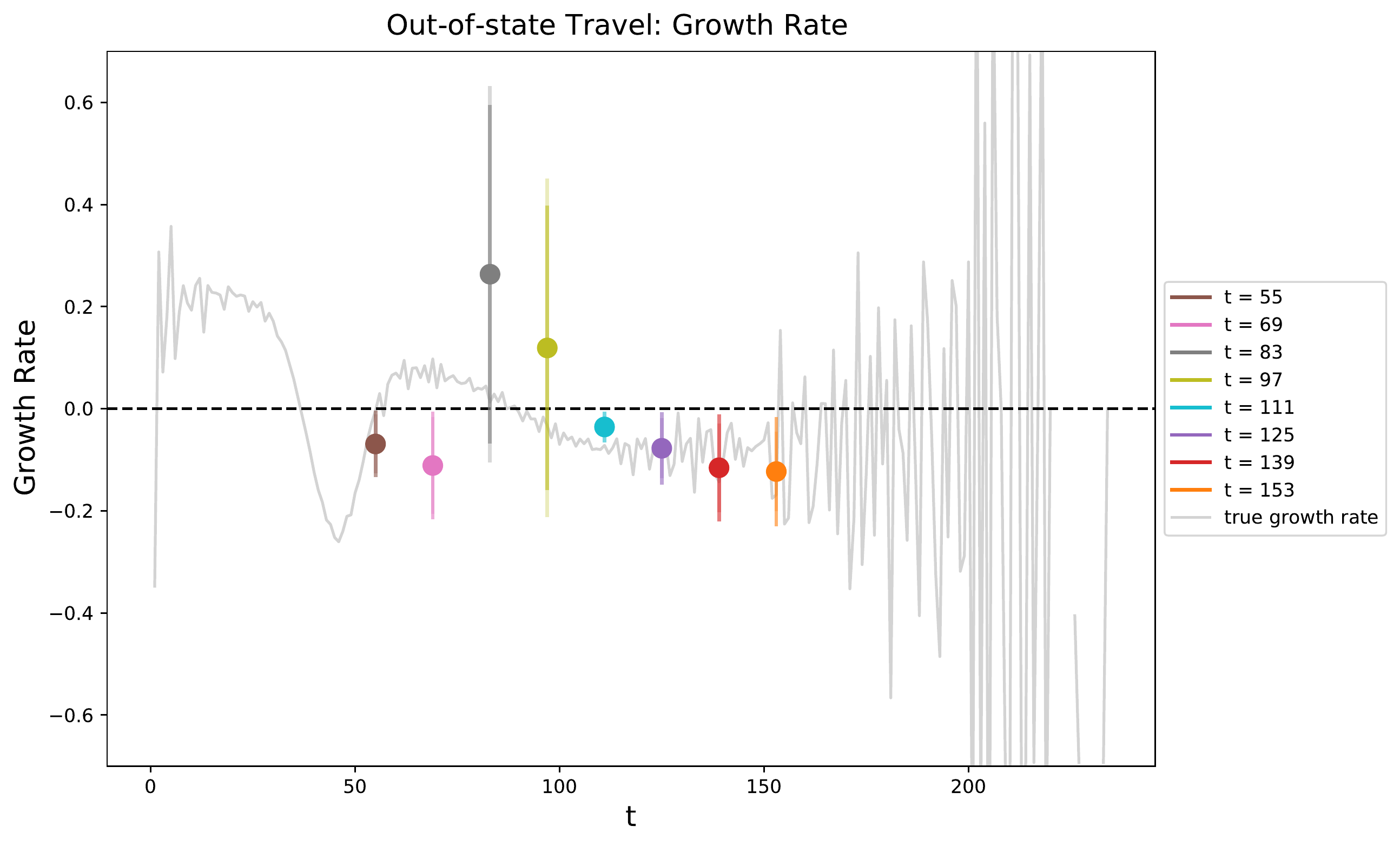}}
    \subfloat(c){\includegraphics[width=0.95\columnwidth]{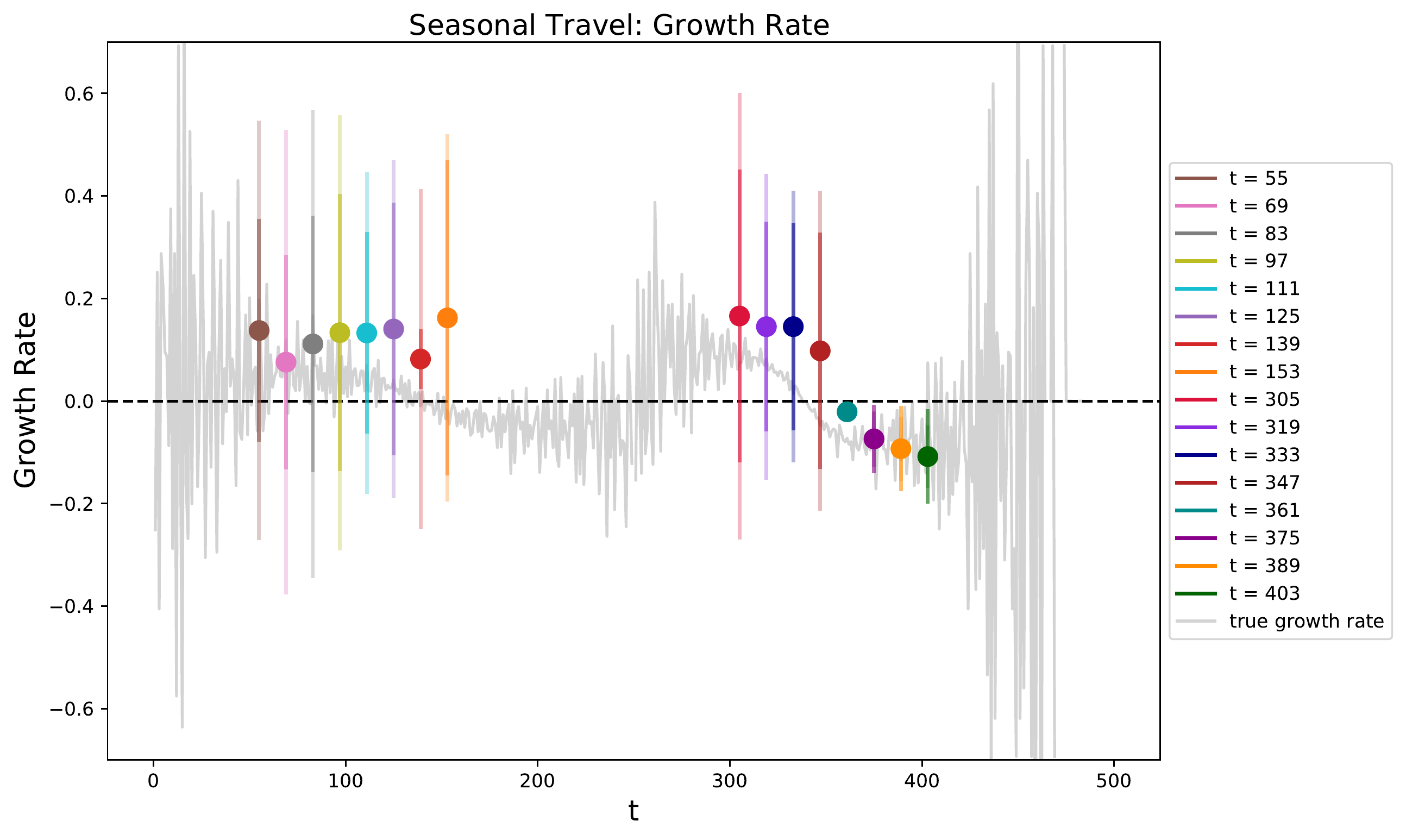}}
    \subfloat(d){\includegraphics[width=0.95\columnwidth]{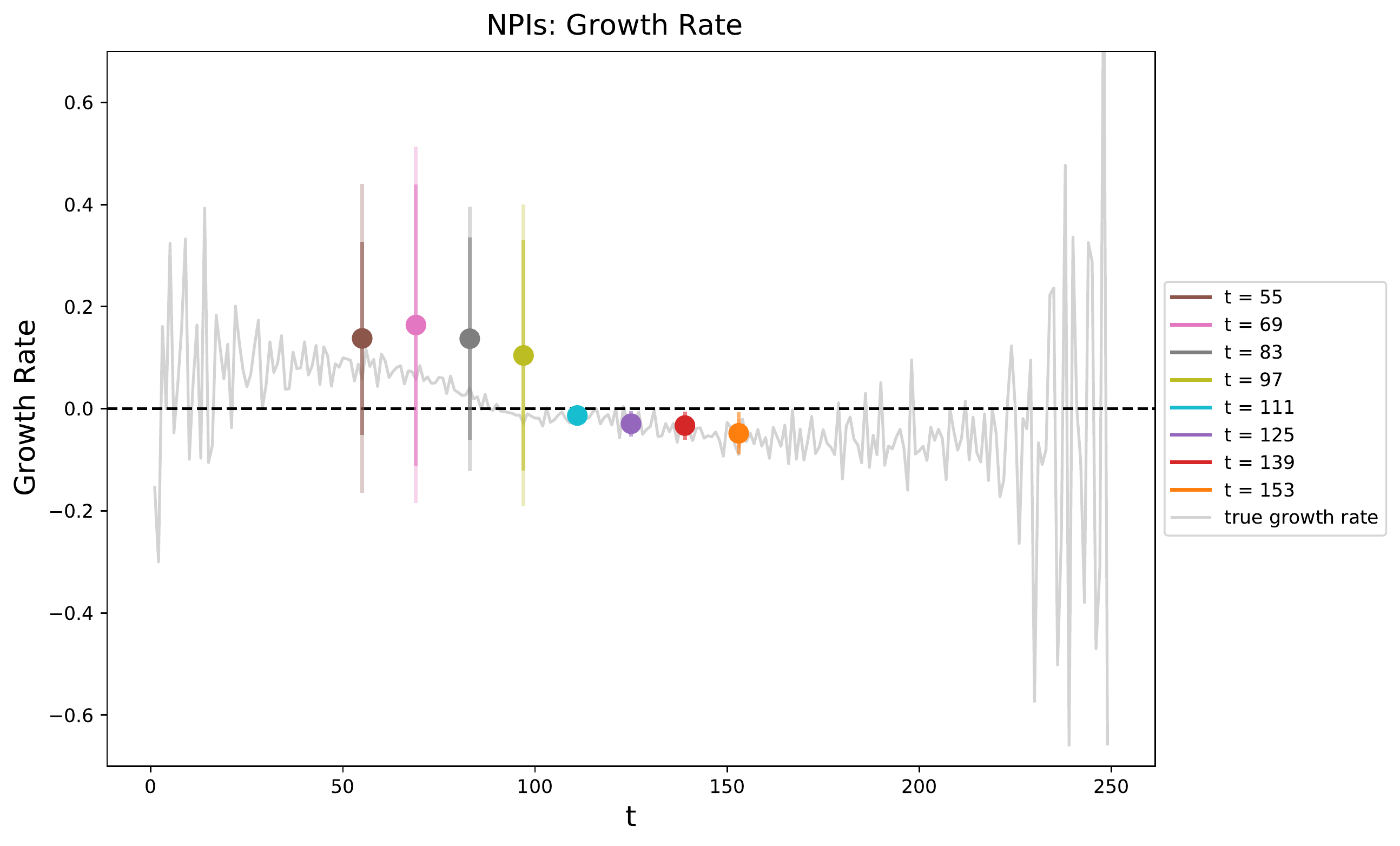}}
    \caption{\textbf{Evaluation of the exponential growth model for each outbreak scenario.} For the \textbf{(a)} baseline, \textbf{(b)} out-of-state travel, \textbf{(c)} seasonal travel, and \textbf{(d)} NPIs scenarios, we show the predicted growth rates (average taken between $t - 35$ and $t$) and the 95\% credible interval for the 8 time points $t$ (dot and vertical line), and their corresponding true growth rates (gray curve). Note that the true growth rates exhibit high variability during time periods with very few new infections (i.e. at the start or end of an outbreak).}
    \label{figs/all_exp_results}
\end{figure*}

The exponential model performs slightly better in the NPIs scenario than in the out-of-state travel scenario (Figures \ref{figs/all_exp_results}d and \ref{figs/scenario5_exp_results}). Although the populations have opposing trajectories at $t \in \{97, 111\}$, the fit of the exponential model falls within our 95\% credible boundary across all time points.

\begin{figure*}
    \centering
    \subfloat(a){\includegraphics[width=0.95\columnwidth]{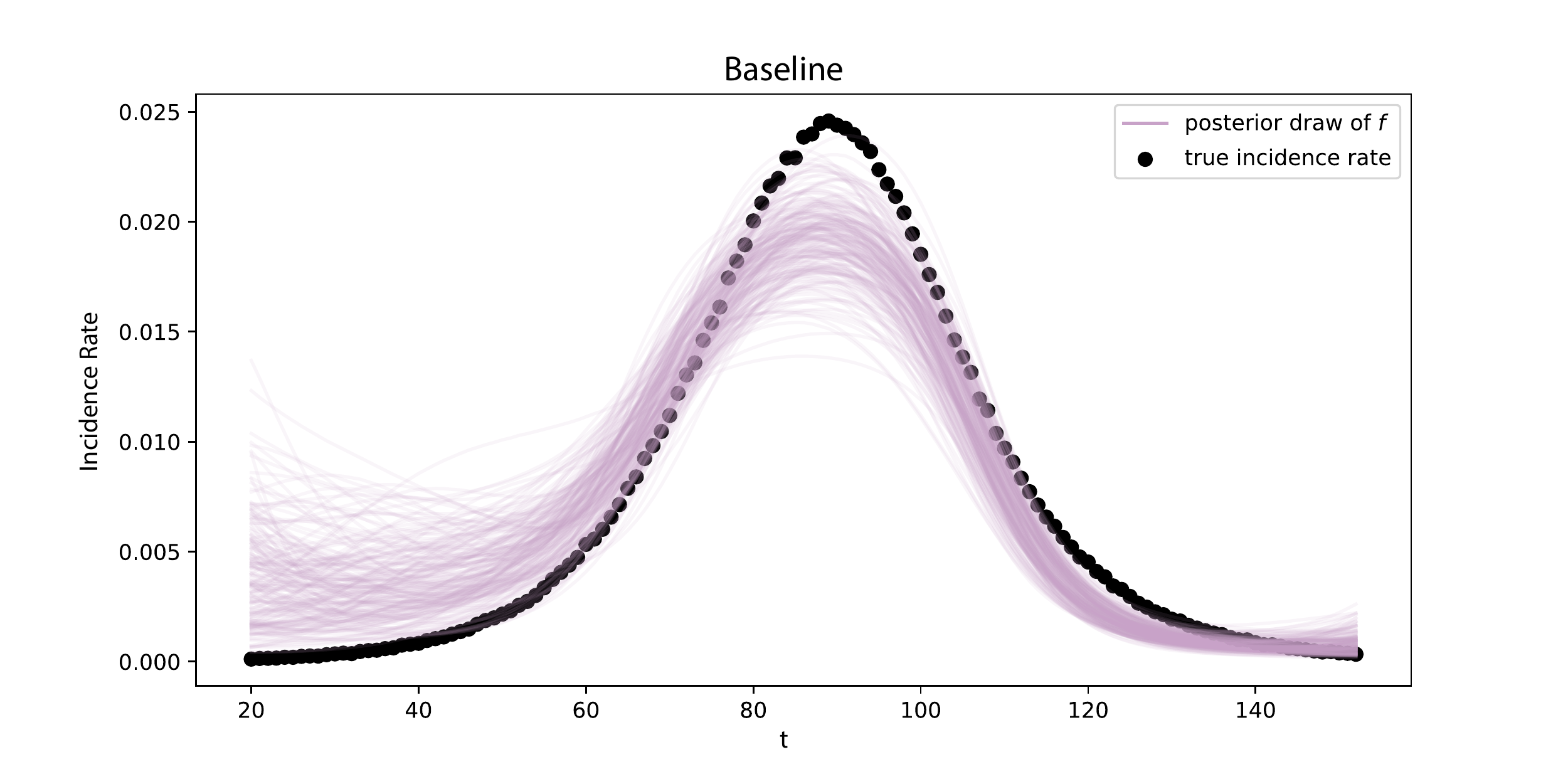}}
    \subfloat(b){\includegraphics[width=0.95\columnwidth]{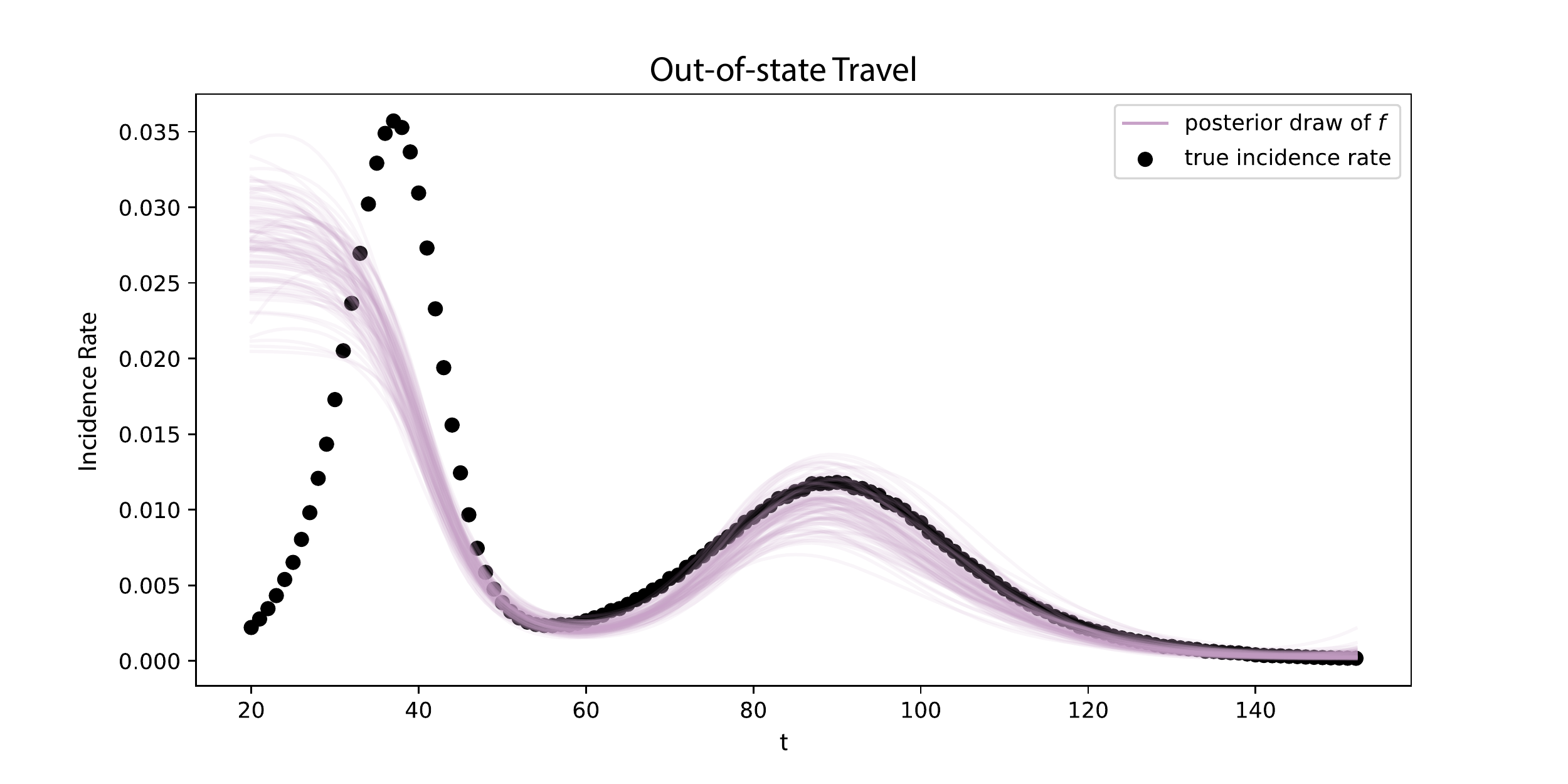}}
    \subfloat(c){\includegraphics[width=0.95\columnwidth]{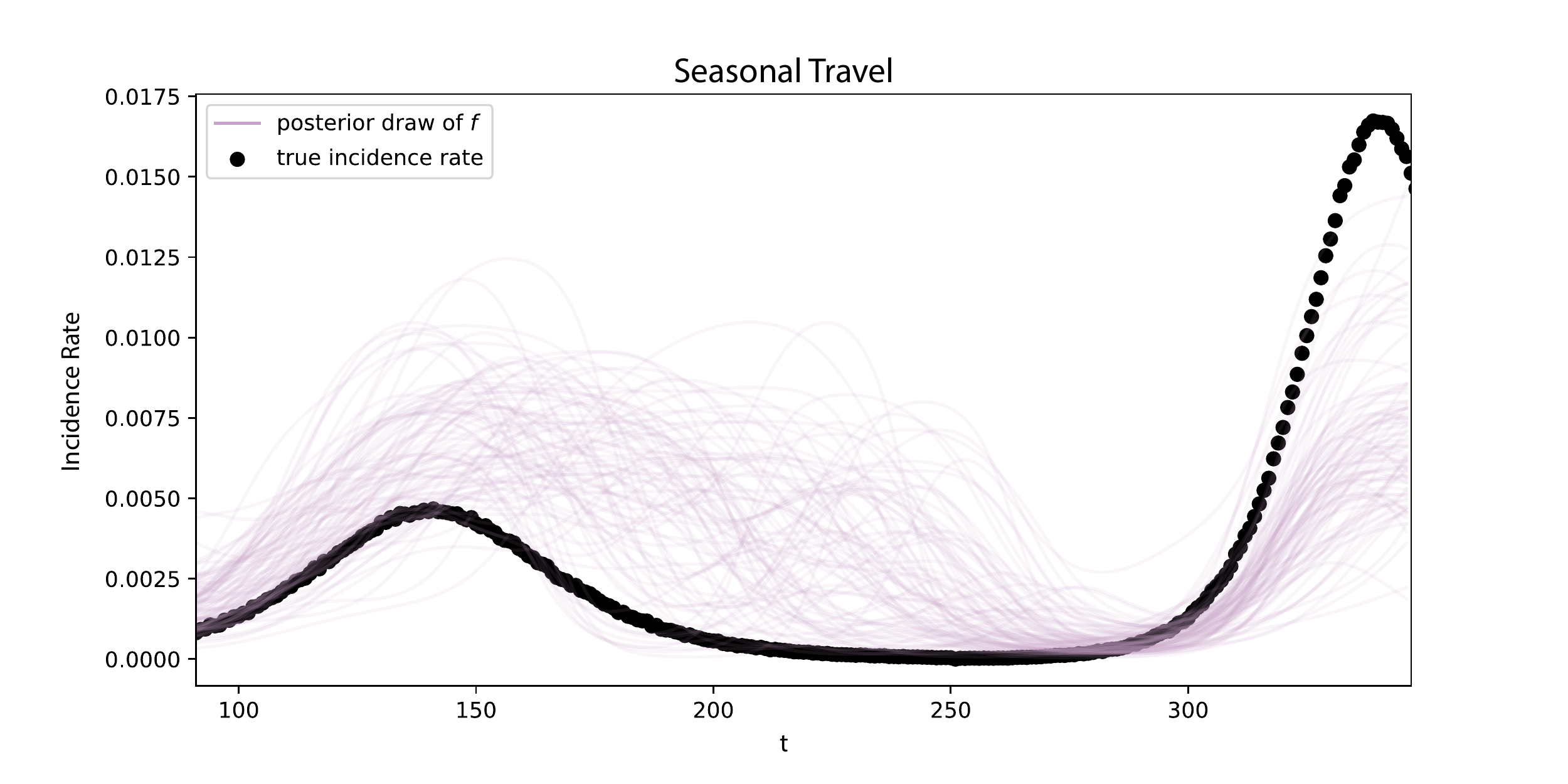}}
    \subfloat(d){\includegraphics[width=0.95\columnwidth]{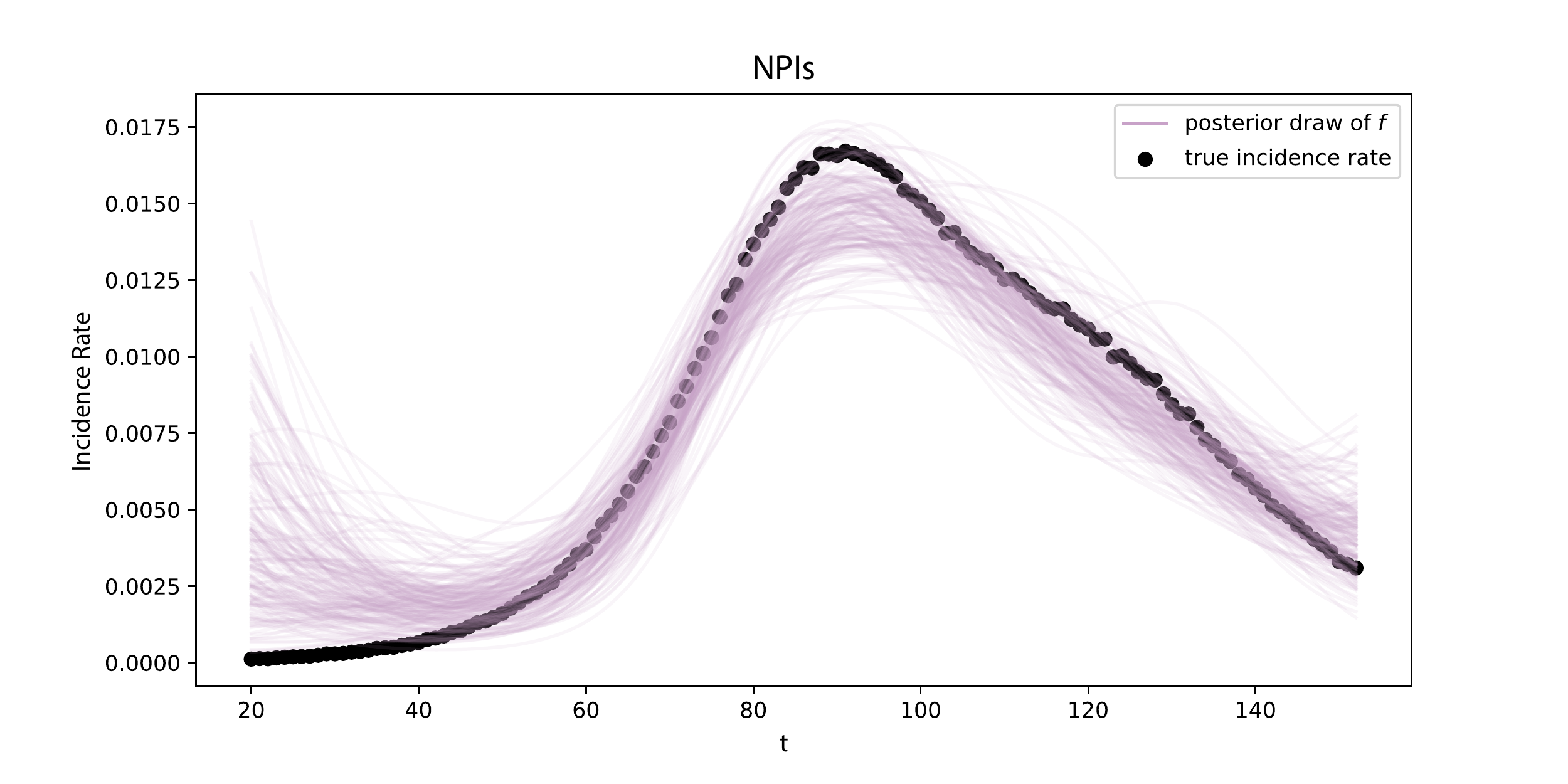}}
    \caption{\textbf{Evaluation of the Gaussian process prior model for each outbreak scenario.} For the \textbf{(a)} baseline, \textbf{(b)} out-of-state travel, \textbf{(c)} seasonal travel, and \textbf{(d)} NPIs scenarios, we show the posterior draws of $t$ (purple) and the true incidence rate in the overall combined population (black).}
    \label{figs/gauss_results}
\end{figure*}

The GP prior model is also able to capture the full trajectory when given enough cross-sections (Figure \ref{figs/gauss_results}d). Similarly with our baseline experiment, the GP prior model struggles when there are too few testing time points (i.e. $t < 50$). However, it is important to note that the GP prior model captures a fair amount of uncertainty for the first cross-section.

%% file: 050conclusion.tex
In this work, we demonstrate the limitations of SoTA epidemiological models in capturing the heterogeneity in underlying aggregated datasets' population structures and epidemiological dynamics. While both the exponential growth and GP prior models are able to infer the direction of the combined epidemic across most time points, the models' estimated trajectories tend to reflect the population and/or epidemic with the highest abundance in surveillance data (e.g.~in the out-of-state travel scenario). Although the exponential growth model seems to accurately estimate simple epidemic trajectories with limited data at time points far from inflection points (e.g.~in the baseline and seasonal travel scenarios), it fails to capture the complexities of two conflicting trajectories (e.g.~in the out-of-state travel scenario). On the other hand, the highly flexible GP prior model can only accurately fit aggregated trajectories when provided sufficient cross-sections that are densely scattered (e.g.~unlike in the seasonal travel scenario).

Future work regarding the epidemiological models includes exploring the effects of model selection (e.g.~kernels $k$, hyperparameters $\eta$ and $\rho$) for the GP prior model in order to address its limitations in estimating aggregated datasets. Our Python implementation of the Ct, exponential growth, and Gaussian process prior models are open source and available. Given the potentially high-impact repercussions of biased and inaccurate models (e.g.~when being used to advise government interventions), we must explore strategies for data aggregations such that models are both robust and fair towards heterogeneous population structures and epidemic dynamics. Moreover, epidemiological models' performance could possibly provide insights on the allocation of testing resources to ensure that adequate data is obtained for fitting the models.

%% file: 060impact.tex
Our work is primarily motivated by the need to account for spatial heterogeneity when estimating epidemiological parameters. This is particularly important due to the use of epidemiological models to inform public health policies (e.g. travel restrictions, mask-wearing) \cite{brauner2021inferring}. Some potential implications of an inaccurate model include (1) inadvertently forcing the population to stay longer than necessary in quarantine or lockdown, which could negatively affect the economy, overall psychological health, and interpersonal relationships; or (2) releasing interventions when incidence is still at the early stage of growth. Nevertheless, our models are unable to make direct recommendations; we are simply highlighting the discrepancies of epidemiological models.

As our results are drawn directly from simulated data, we must consider the advantages and disadvantages of generating our own data rather than using real-world data. One major advantage is that it enables us to preserve the privacy of individuals \cite{el2021evaluating}. Moreover, because it is relatively straightforward to manipulate the parameters of compartmental models (e.g. SEIR model), we can simulate a wide range of desired scenarios for investigation. While many studies have validated synthetic data as adequate proxies for real-world data \cite{el2021evaluating}, there may still be artificial biases introduced by the simulations.

Finally, our criticism of epidemiological models should be considered in the context of current state-of-the-art models' capabilities. Although standard models often make oversimplifying assumptions, it is important to recognize that modeling real world dynamics is highly complex and challenging. The ability to capture even some of the key contributing factors of an epidemic using a simple model is valuable, as it can enable further extrapolation to more robust models.

%% file: 070appendix.tex
\setcounter{figure}{0} 
\renewcommand{\thefigure}{S\arabic{figure}} 
\setcounter{table}{0} 
\renewcommand\thetable{S\arabic{table}} 

\onecolumn

\begin{appendices}

\section{Inference Models}

The inference models from \cite{hay2020estimating} can be decomposed into two components: (1) a custom Ct likelihood and (2) a prior on the growth rate parameter(s). The custom likelihood for a single-cross section at time $t$ is as follows:
\begin{equation*}
    \mathcal{L}(X_1,...,X_n | \pi_{t - A_{max}},...,\pi_{t - a})
= \prod_{i = 1}^n \bigg[ \bigg(\sum_{a = 1}^{A_{max}} p_a(X_i)\phi_a \pi_{t - a}\bigg)^{\mathbf{I}(X_i \leq C_{LOD})}
\bigg( 1 - \sum_{a = 1}^{A_{max}}\phi_a\pi_{t - a}\bigg)^{\mathbf{I}(X_i > C_{LOD})} \bigg]
\end{equation*}
where, for testing day $t$, $\pi_{t - a} := $ probability that a randomly-selected individual in the population was infected on day $t - a$, $p_a(x) :=$ probability that Ct value is $x$ for a test conducted $a$ days after infection, and $\phi_a :=$ probability of a Ct value being detectable $a$ days after infection. Since we only consider detectable Ct values for a cross-section, the likelihood simplifies into:
\begin{equation*}
    \mathcal{L}(X_1,...,X_n | \pi_{t - A_{max}},...,\pi_{t - a}) = \frac{\prod_{i = 1}^n (\sum_{a = 1}^{A_{max}} p_a(X_i)\phi_a\pi_{t - a}])}{( \sum_{a = 1}^{A_{max}}\phi_a\pi_{t - a}])^n}    
\end{equation*}
where $n$ is the number of detectable Ct values in the cross-section. Since the likelihood can lead to overflow values, we use log-likelihood. Using the log-likelihood does not disrupt the optimization problem because log is monotonic.
\begin{align*}
    & \implies l(X_1,..., X_n | \pi_{t - A_{max}},...,\pi_{t - a}) = \log \mathcal{L} \\
    & = \sum_{i = 1}^n \log (\sum_{a = 1}^{A_{max}} p_a(X_i)\phi_a\pi_{t - a}) - n \log( \sum_{a = 1}^{A_{max}}\phi_a\pi_{t - a})
\end{align*}

\subsection{Single Cross-Section Model}
The single-cross section model assigns a normal prior on a single growth rate parameter $\beta \sim \mathcal{N}(0, 0.25)$. We assume a fixed growth rate for $A_{max} = 35$ days. Daily incidence within that interval is $\pi_{t - a} = \exp(\beta(t - a))$. With both components, we obtain a posterior distribution:
\begin{equation*}
    p(\beta | X_1,...,X_n) \propto l(X_1,..., X_n | \beta)p(\beta)
\end{equation*}
We sample from the posterior and obtain draws of $\beta$. We then use these $\beta$'s to construct a posterior predictive, which we compare against the true incidence rates. Note that when we compare against incidence \textit{rates}, we need to modify the likelihood function such that $\pi = [\pi_{t - a} / \sum_{a = 1}^{A_{max}}\pi_{t - a}]$ reflects an incidence rate.

\subsection{Multiple Cross-Section Model}

For the multiple cross section model, we leverage the function space view of Gaussian process priors \cite{rasmussen2003gaussian}. Instead of sampling a parameter, we directly sample functions that represent the incidence rate curve. A GP prior is defined entirely by its kernel function, which are usually specified directly by domain experts. Accounting for multiple cross sections is a natural extension from the single cross section model and is given by:
\begin{equation*}
    \mathcal{L}(X_1^{t_1},...,X_{n_1}^{t_1}, ..., X_1^{t_T},...,X_{n_T}^{t_T} | \pi_{t - A_{max}^{(T)}},...,\pi_{t_T - 1}) = \prod_{j = 1}^T \bigg [\frac{\prod_{i = 1}^{n_j} (\sum_{a = 1}^{A_{max}^{(j)}} p_a(X_i^{t_j})\phi_a\pi_{t_j - a})}{( \sum_{a = 1}^{A_{max}^{(j)}}\phi_a\pi_{t_j - a})^{n_j}} \bigg]    
\end{equation*}
The log likelihood is derived below:
\begin{equation*}
    \log \mathcal{L}(X_1^{t_1},...,X_{n_1}^{t_1}, ..., X_1^{t_T},...,X_{n_T}^{t_T} | \pi_{t - A_{max}^{(T)}},...,\pi_{t_T - 1})
= \sum_{j = 1}^T \bigg[ \sum_{i = 1}^{n_j} \bigg[ \log [(\sum_{a = 1}^{A_{max}^{(j)}} p_a(X_i^{t_j})\phi_a\pi_{t_j - a})] - {n_j}\log ( \sum_{a = 1}^{A_{max}^{(j)}}\phi_a\pi_{t_j - a}) \bigg] \bigg]    
\end{equation*}
Again, we need to modify the likelihood function to replace $\pi = [\pi_{t - a} / \sum_{a = 1}^{A_{max}}\pi_{t - a}]$ such that it generates incidence rates.

\newpage

\section{Supplementary Tables}

\begin{table}[!h]
  \caption{\textbf{Testing times and counts from Ct simulations.} We present the number of Ct values collected at each testing time for all scenarios.}
  \label{tab:ct_values}
  \centering
  \begin{tabular}{l c c c c c c c c c}
    \toprule
    Scenario & $t = 55$ & $t=69$ & $t=83$ & $t=97$ & $t=111$ & $t=125$ & $t=139$ & $t=153$ \\
    \midrule
    Baseline & 49 & 172 & 466 & 777 & 612 & 342 & 140 & 50 \\
    Out-of-state travel & 1148 & 466 & 523 & 710 & 616 & 334 & 140 & 61\\
    Seasonal travel & 6 & 9 & 27 & 71 & 120 & 240 & 402 & 463 \\
    NPIs & 52 & 221 & 553 & 883 & 959 & 787 & 573 & 390\\
    \toprule
     & $t=305$ & $t=319$ & $t=333$ & $t=347$ & $t=361$ & $t=375$ & $t=389$ & $t=403$\\
        \midrule
    Seasonal travel & 36 & 149 & 412 & 737 & 619 & 328 & 142 & 50 \\
  \bottomrule
\end{tabular}
\end{table}

\section{Supplementary Figures}

\begin{figure*}[!h]
    \centering
    \subfloat(a){\includegraphics[width=0.4\textwidth]{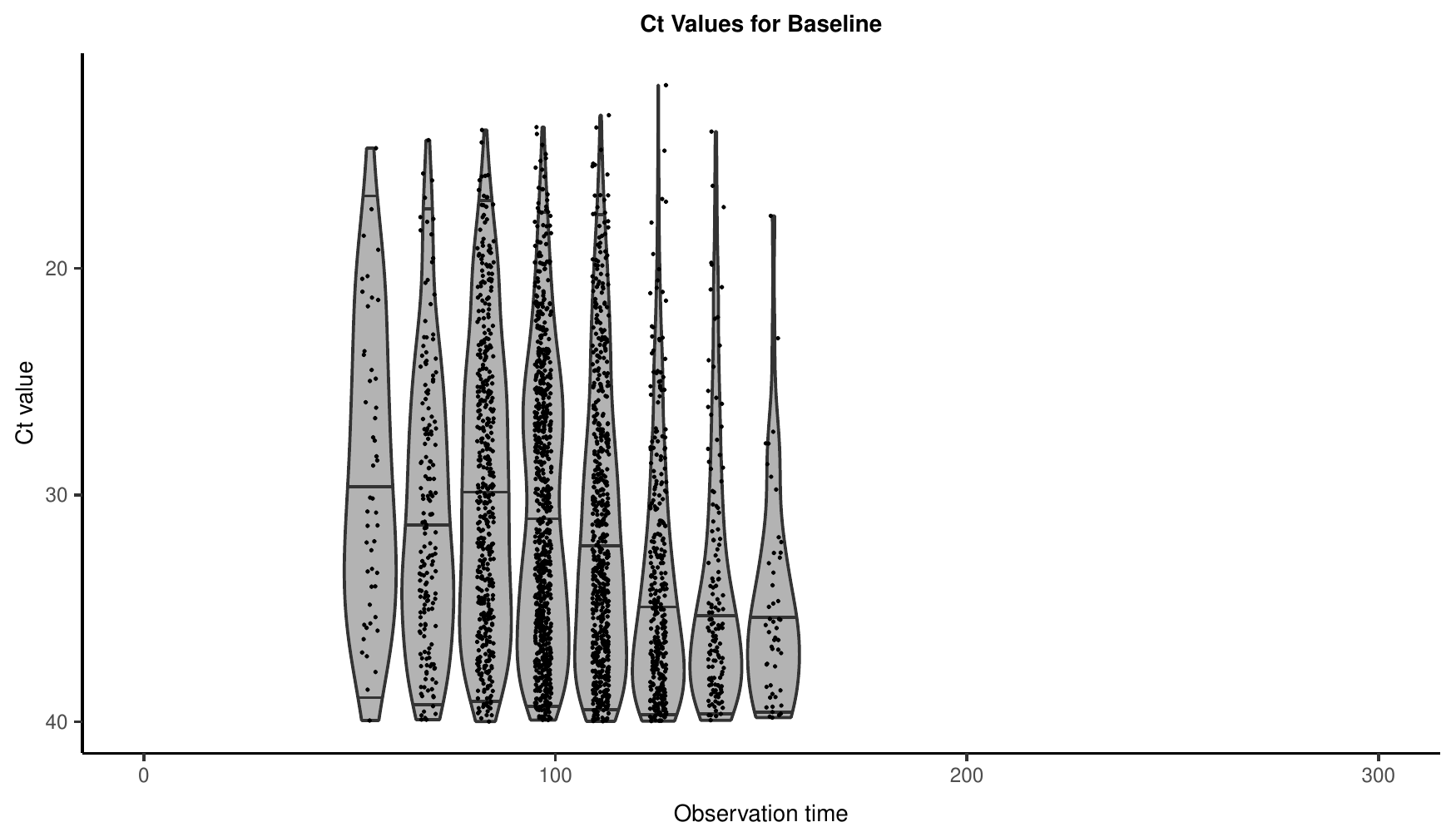}}
    \subfloat(b){\includegraphics[width=0.4\textwidth]{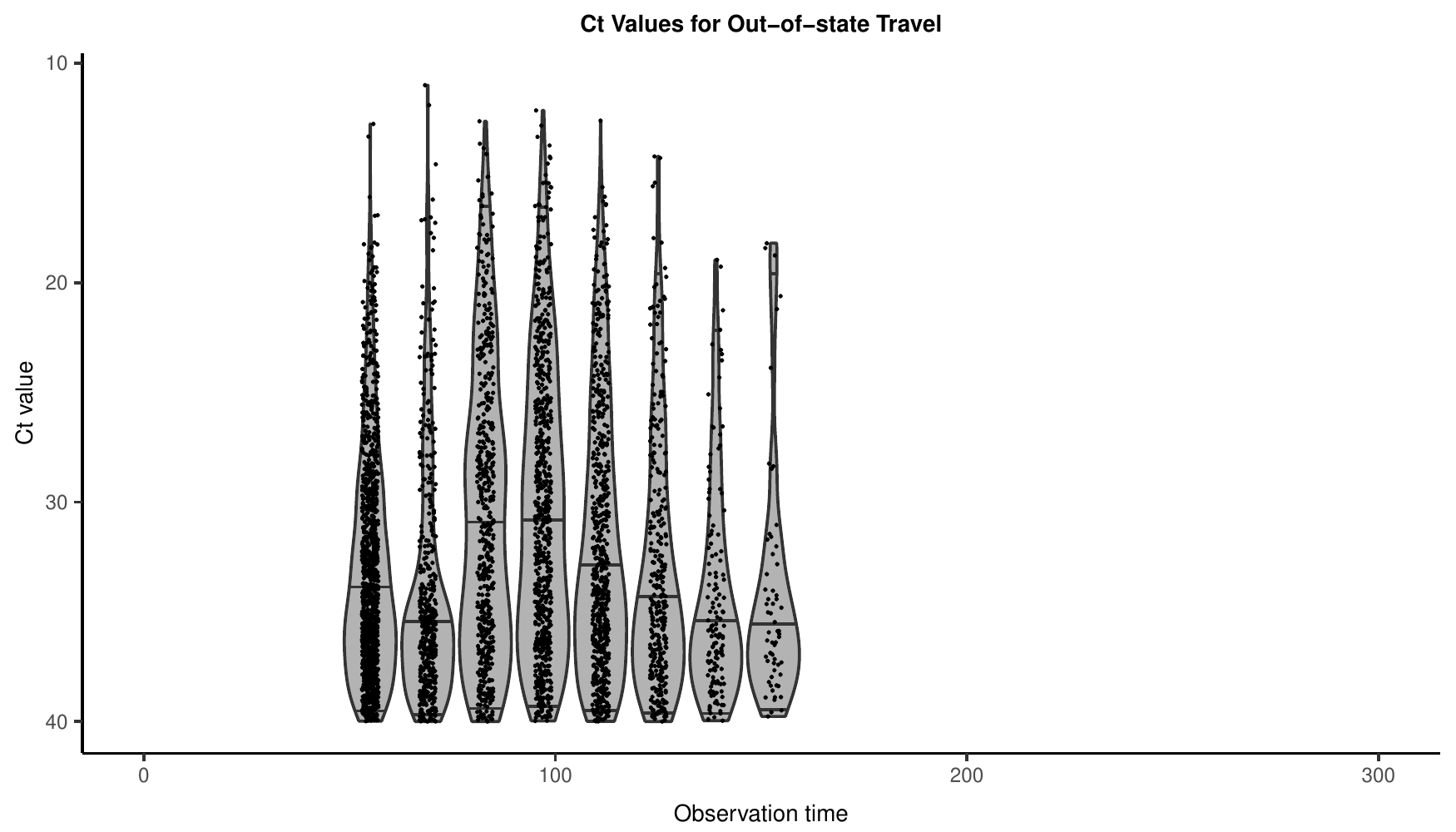}}
    \subfloat(c){\includegraphics[width=0.4\textwidth]{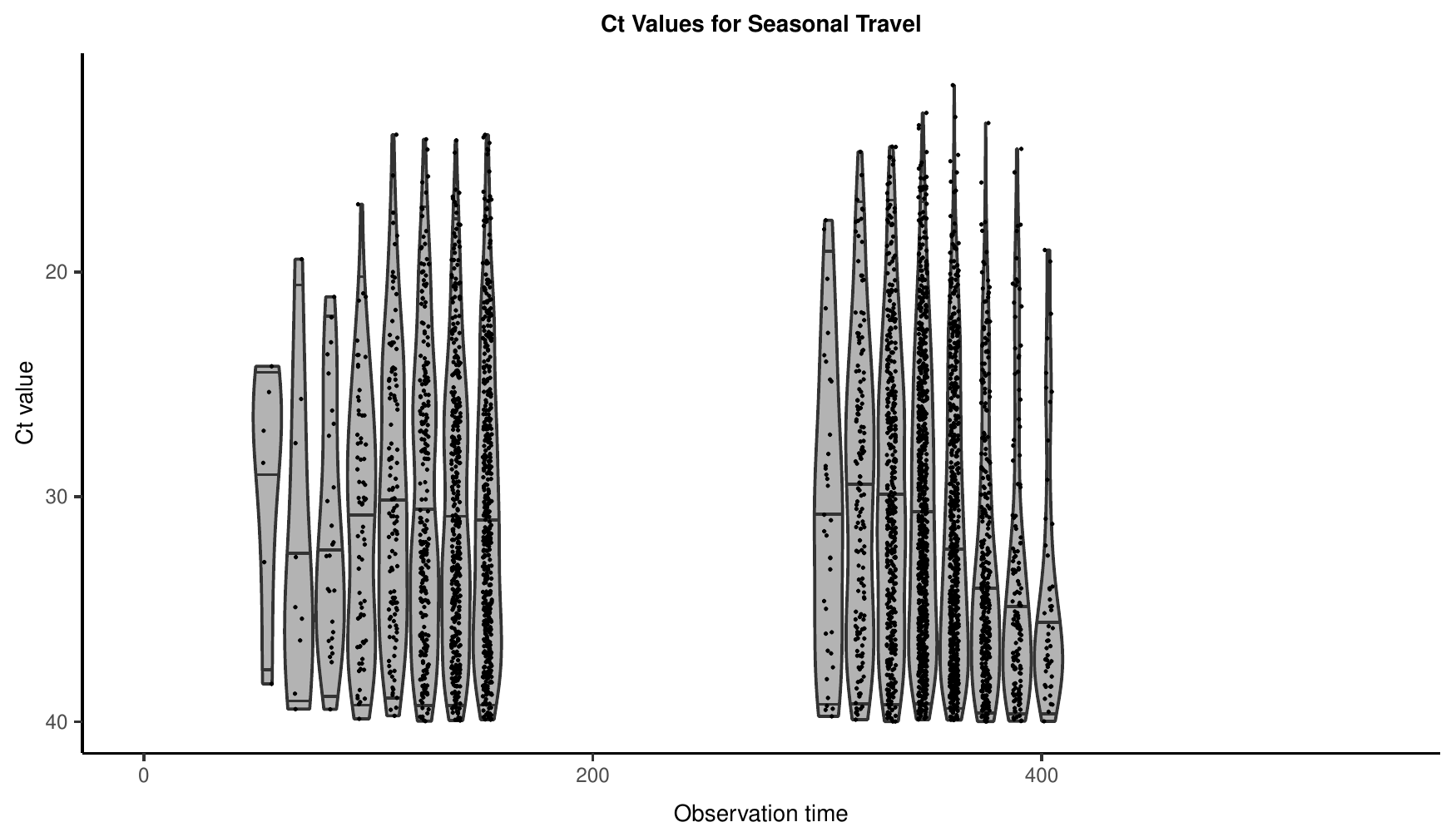}}
    \subfloat(d){\includegraphics[width=0.4\textwidth]{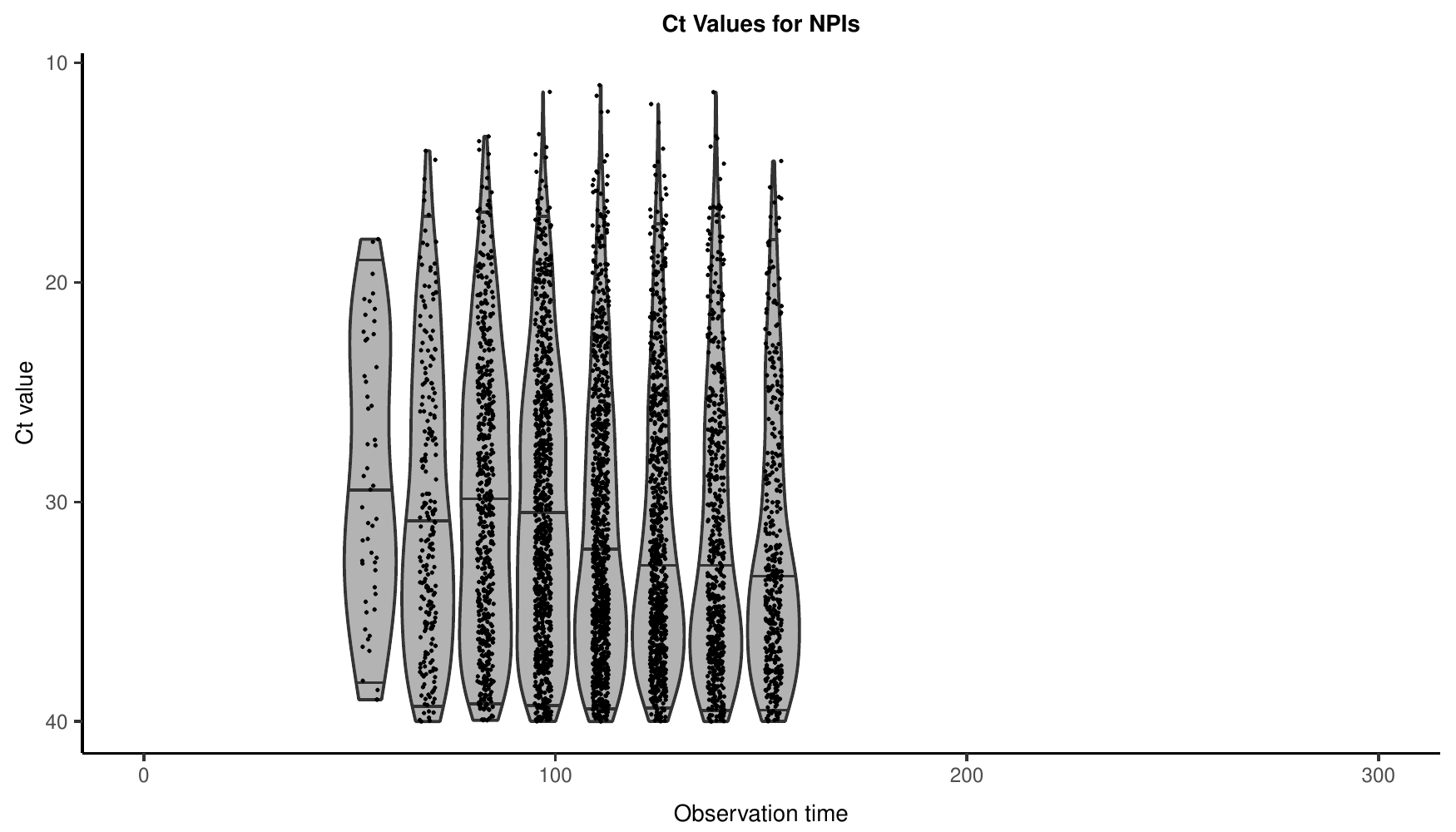}}
    \caption{\textbf{Ct values of simulated data.} For the \textbf{(a)} baseline, \textbf{(b)} out-of-state travel, \textbf{(c)} seasonal travel, and \textbf{(d)} NPIs scenarios, we show the distribution of Ct values across the combined populations at each observed time point.}
    \label{figs/sim_ct_violin}
\end{figure*}

\begin{figure*}[!h]
    \centering
    \includegraphics[width=\textwidth]{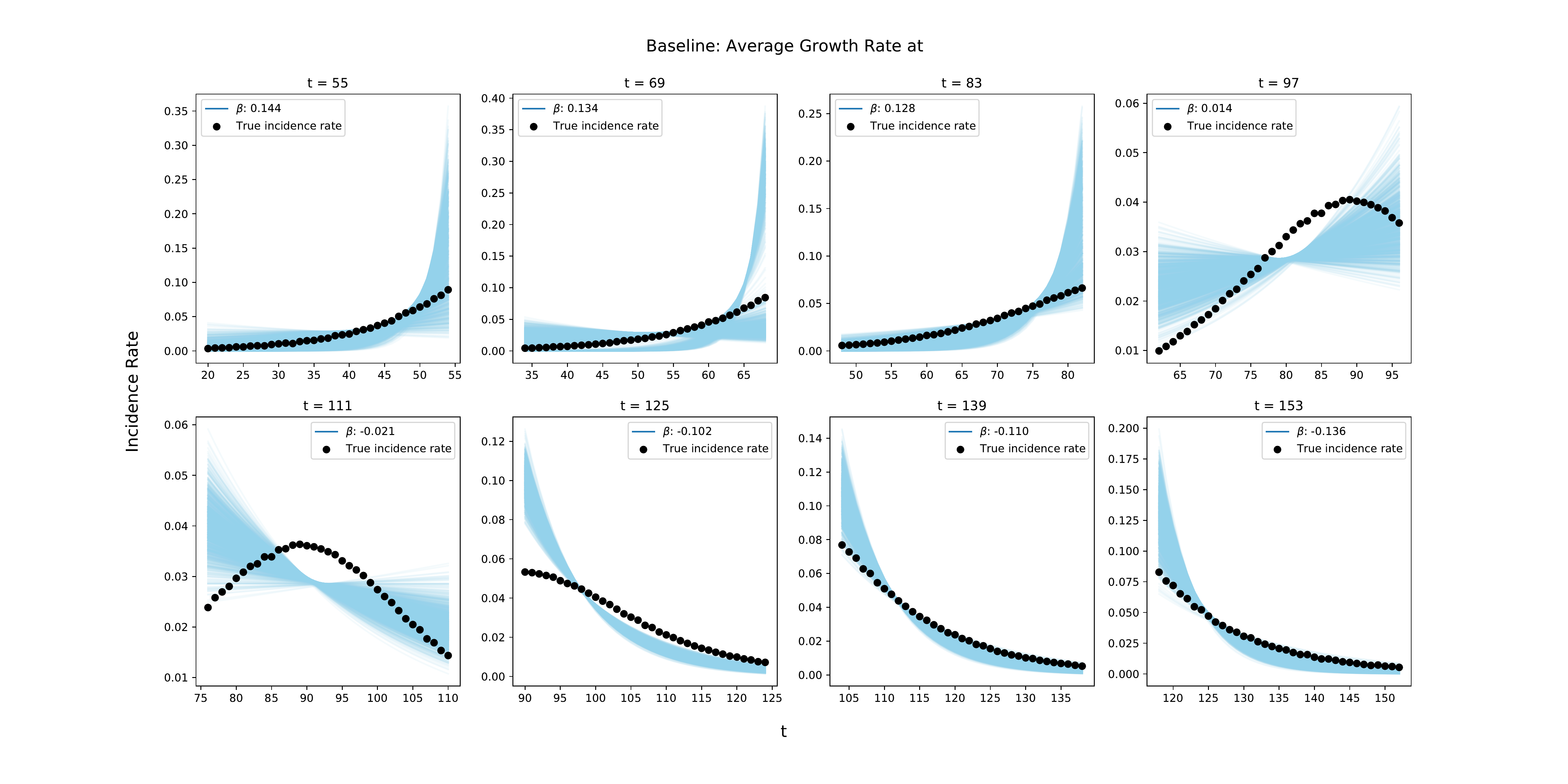}
    \caption{\textbf{Evaluation of the exponential growth model on the baseline scenario.} For each of the 8 time points, we show the predicted growth rates $\beta$ from the exponential growth model (blue) and the true incidence rate (black). The value for $\beta$ shown in each figure legend is the posterior mean estimate. The model is able to estimate the correct direction of the epidemic; however, despite the simple trajectory, it has difficulty predicting growth rates near the inflection point of the curve.}
    \label{figs/scenario0_exp_results}
\end{figure*}

\begin{figure*}[!h]
    \centering
    \includegraphics[width=\textwidth]{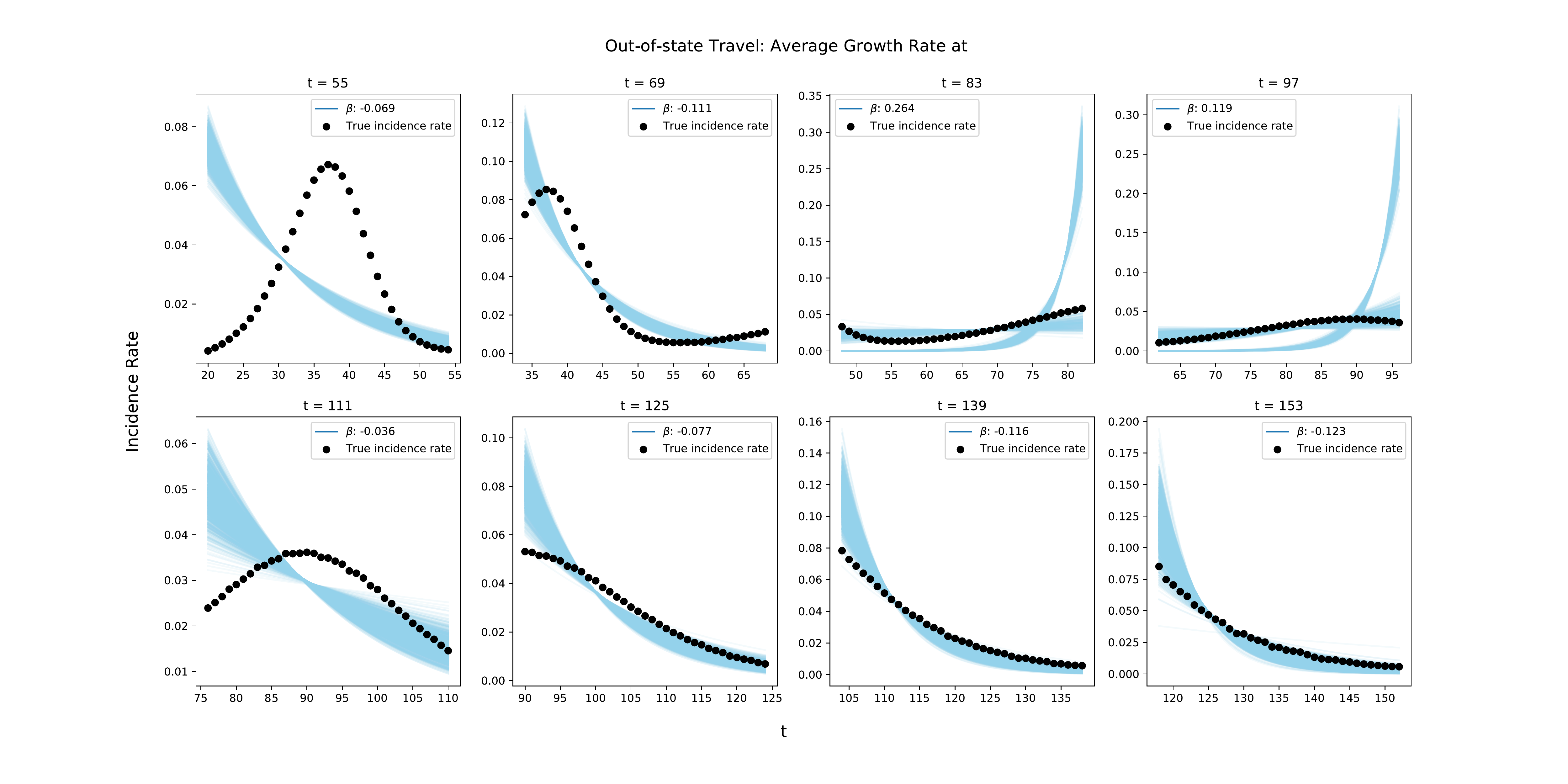}
    \caption{\textbf{Evaluation of the exponential growth model on the out-of-state travel scenario.} For each of the 8 time points, we show the predicted growth rates $\beta$ from the exponential growth model (blue) and the true incidence rate (black). The value for $\beta$ shown in each figure legend is the posterior mean estimate. The model is able to correctly fit the combined incidence curve within all testing times except at $t \in \{55, 69, 111\}$, of which $t \in \{55, 69\}$ are locations of high variability across the populations, and $t = 111$ is an inflection point.}
    \label{figs/scenario1_exp_results}
\end{figure*}

\begin{figure*}[!h]
    \centering
    \includegraphics[width=\textwidth]{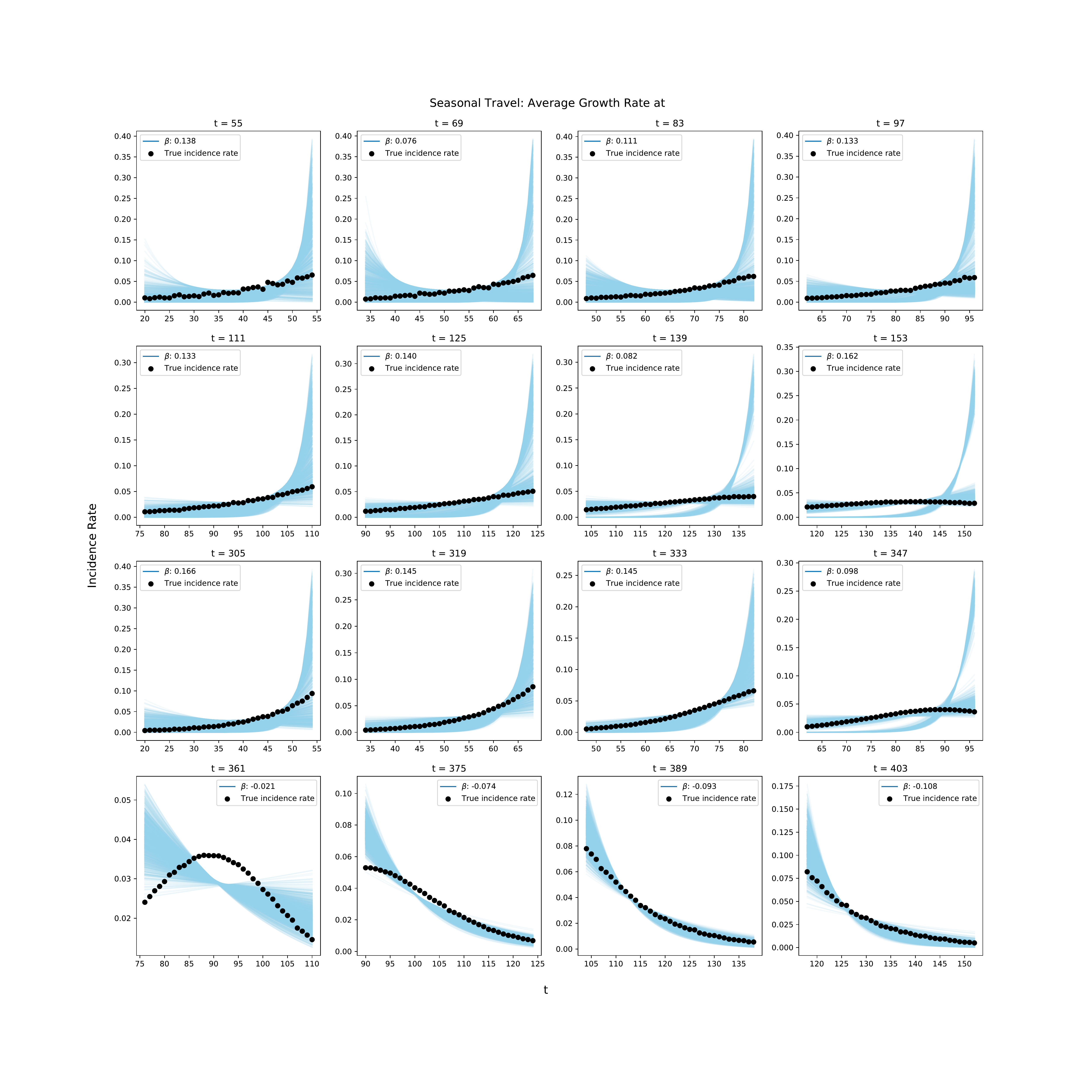}
    \caption{\textbf{Evaluation of the exponential growth model on the seasonal travel scenario.} For each of the 16 time points, we show the predicted growth rates $\beta$ from the exponential growth model (blue) and the true incidence rate (black). The value for $\beta$ shown in each figure legend is the posterior mean estimate. As there is no overlap between the two populations in this scenario, we observe very little variability in the aggregated dataset. The exponential growth model is robust in all sampled times except at $t = 361$, the inflection point of the curve.}
    \label{figs/scenario2_exp_results}
\end{figure*}

\begin{figure*}[!h]
    \centering
    \includegraphics[width=\textwidth]{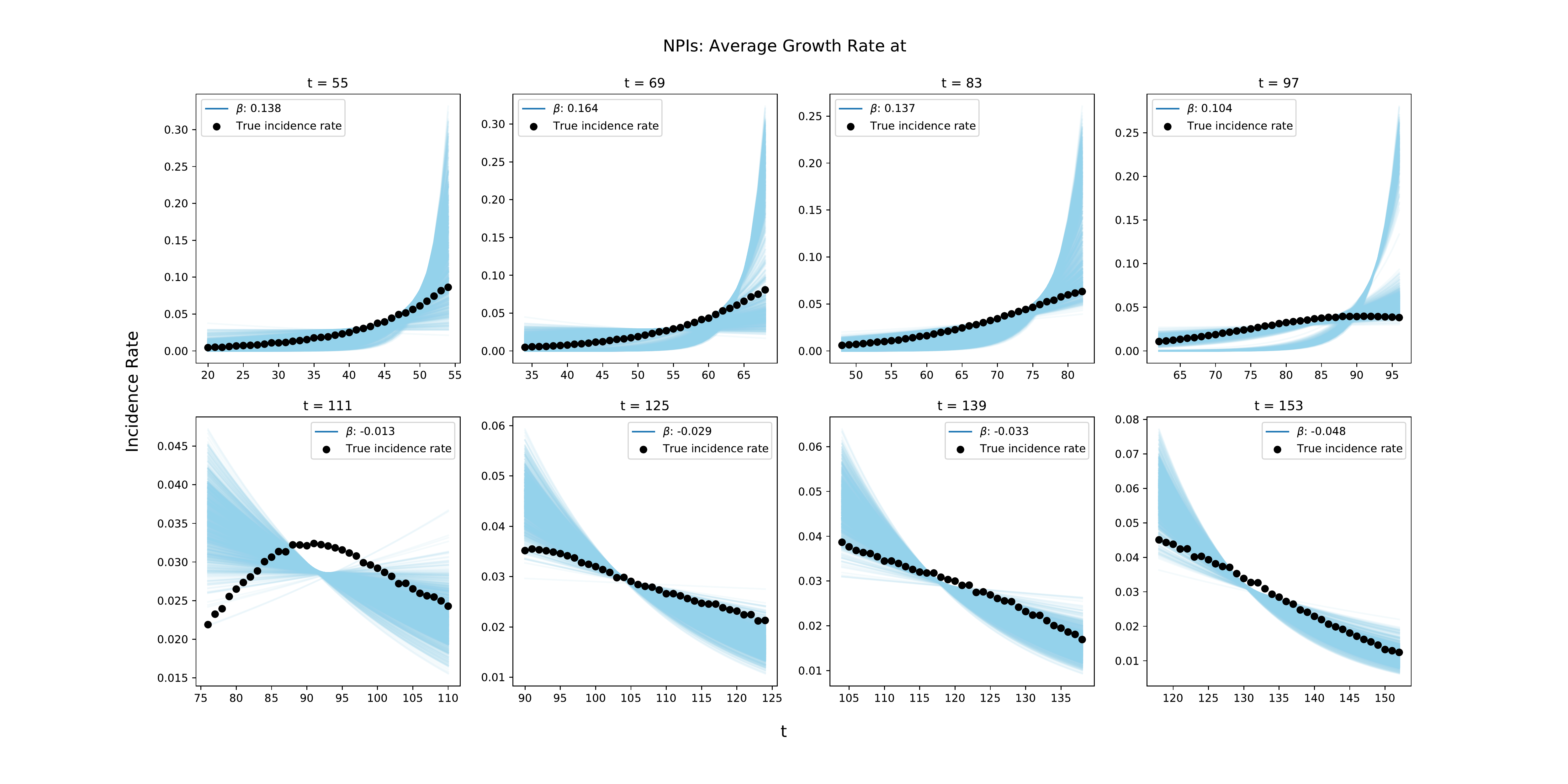}
    \caption{\textbf{Evaluation of the exponential growth model on the NPIs scenario.} For each of the 8 time points, we show the predicted growth rates $\beta$ from the exponential growth model (blue) and the true incidence rate (black). The value for $\beta$ shown in each figure legend is the posterior mean estimate. The model performs relatively well in this scenario. Although the populations have opposing trajectories at $t \in \{97, 111\}$, the fit of the exponential model falls within the 95\% credible interval across all time points.}
    \label{figs/scenario5_exp_results}
\end{figure*}

\end{appendices}